\documentclass[journal]{IEEEtran}
\usepackage{times}
\usepackage{epsfig}
\usepackage{graphicx}
\usepackage{amsmath}
\usepackage{amssymb}
\usepackage{multirow}
\usepackage{array}
\usepackage{ulem}
\usepackage{cancel}
\usepackage{color}
\usepackage{cite}

\ifCLASSINFOpdf
\else
\fi
\begin{document}

\indent

© 2020 IEEE. Personal use of this material is permitted. Permission from IEEE must be obtained for all other uses, in any current or future media, including reprinting/republishing this material for advertising or promotional purposes, creating new collective works, for resale or redistribution to servers or lists, or reuse of any copyrighted component of this work in other works. Digital Object Identifier 10.1109/TCSVT.2020.3037688

%
\title{SCGAN: Saliency Map-guided Colorization with Generative Adversarial Network}
%
%
%

\author{Yuzhi~Zhao,~\IEEEmembership{Student~Member,~IEEE,}
        Lai-Man~Po,~\IEEEmembership{Senior~Member,~IEEE,}
        Kwok-Wai~Cheung,~\IEEEmembership{Member,~IEEE,}
        Wing-Yin~Yu,
        Yasar Abbas Ur Rehman,~\IEEEmembership{Member,~IEEE}
\thanks{Manuscript received April 15, 2020; revised August 11, 2020, October 3, 2020 and November 5, 2020; accepted November 8, 2020. This work was supported by an Internal Funds Scheme from City University of Hong Kong under Project 9678141. This article was recommended by Associate Editor H. Xiong. \textit{(Corresponding author: Yuzhi Zhao.)}}
\thanks{Y. Zhao, L.-M. Po, and W.-Y. Yu are with the Department of Electronic Engineering, City University of Hong Kong, Hong Kong (e-mail: yzzhao2-c@my.cityu.edu.hk; eelmpo@cityu.edu.hk; wingyinyu8-c@my.cityu.edu.hk).}
\thanks{K.-W. Cheung is with School of Communication, the Hang Seng University of Hong Kong, Hong Kong (e-mail: keithcheung@hsmc.edu.hk).}
\thanks{Y.-A.-U. Rehman is with TCL Corporate Research Hong Kong, Hong Kong (e-mail: yasar.abbas@my.cityu.edu.hk).}
\thanks{This article has supplementary material provided by the authors and color versions of one or more figures available at https://doi.org/10.1109/TCSVT.2020.3037688.}
\thanks{Digital Object Identifier 10.1109/TCSVT.2020.3037688}
}

%
%

\markboth{IEEE Transactions on Circuits and Systems for Video Technology}%
{Shell \MakeLowercase{\textit{Zhao et al.}}: SCGAN: Saliency Map-guided Colorization with Generative Adversarial Network}
%



\maketitle

\begin{abstract}

Given a grayscale photograph, the colorization system estimates a visually plausible colorful image. Conventional methods often use semantics to colorize grayscale images. However, in these methods, only classification semantic information is embedded, resulting in semantic confusion and color bleeding in the final colorized image. To address these issues, we propose a fully automatic Saliency Map-guided Colorization with Generative Adversarial Network (SCGAN) framework. It jointly predicts the colorization and saliency map to minimize semantic confusion and color bleeding in the colorized image. Since the global features from pre-trained VGG-16-Gray network are embedded to the colorization encoder, the proposed SCGAN can be trained with much less data than state-of-the-art methods to achieve perceptually reasonable colorization. In addition, we propose a novel saliency map-based guidance method. Branches of the colorization decoder are used to predict the saliency map as a proxy target. Moreover, two hierarchical discriminators are utilized for the generated colorization and saliency map, respectively, in order to strengthen visual perception performance. The proposed system is evaluated on ImageNet validation set. Experimental results show that SCGAN can generate more reasonable colorized images than state-of-the-art techniques.


\end{abstract}


%
\IEEEpeerreviewmaketitle

\section{Introduction}
%
%
%
%

\IEEEPARstart{I}{MAGE} colorization is the process of assigning plausible and perceptual colors to each pixel in the input image. It has found a wide array of applications in computer vision, such as multispectral image colorization \cite{nyberg2018unpaired, berg2018generating}, image compression \cite{baig2017multiple}, cartoon colorization \cite{zhang2018two, qu2006manga}, restoration of old photographs and films\cite{chen2018automatic}, fake colorization detection \cite{guo2018fake} and even assisting other tasks like classification and segmentation \cite{larsson2017colorization}. However, without prior information on the colors of the objects in the input intensity image, the colorization results may vary largely from system to system. Notably, the semantic confusion (which color should be assigned to each object in the image), color bleeding (spreading of colors beyond the object boundary), edge distortion, and object intervention are some key problems in the current automatic image colorization tasks.

There are multiple possible colors for an object in the image. Assigning a proper color to the object in an image is still an open research problem in multiple domains. In recent decades, a multitude of algorithms have been proposed to solve this problem. These algorithms can be divided into three possible categories: (1) Scribble-based methods \cite{levin2004colorization, huang2005adaptive, yatziv2006fast, xu2013sparse, xu2009efficient, zhang2017real, chen2012manifold, luan2007natural, sheng2013video}, (2) example-based methods \cite{bugeau2013variational, li2017example, reinhard2001color, fang2019superpixel, charpiat2008automatic, he2018deep, gupta2012image, ironi2005colorization, welsh2002transferring, tai2005local, iizuka2019deepremaster, chia2011semantic, liu2008intrinsic}, and (3) fully-automatic methods \cite{deshpande2017learning, royer2017probabilistic, larsson2016learning, zhao2018pixel, zhao2019pixelated, isola2017image, zhang2016colorful, iizuka2016let, cao2017unsupervised, cheng2015deep, deshpande2015learning, guadarrama2017pixcolor, lei2019fully, vitoria2020chromagan, DeOldify}. The first two categories of algorithms require human interactions for assigning reasonable colors to various objects in the input-intensity image. As a result, these algorithms are highly correlated with the rationality of the human hints, which makes them labor-intensive and less robust to errors. For example, the scribble-based methods utilize the color hints, provided by the user, to assign different colors to the objects in the image. Similarly, the example-based methods require an additional color image to infer the chrominance intensity of different objects in the input image.

On the other hand, fully automatic approaches utilize end-to-end learning to directly learn the relationship between an input grayscale image and the corresponding color embeddings, without any human intervention. Most of these approaches utilize the deep Convolutional Neural Networks (CNN). Normally they are trained on large-scale datasets such as ImageNet \cite{russakovsky2015imagenet} (1.3M images) and Places \cite{zhou2017places} (1.8M images) to encode the semantic information for image colorization. For instance, Larsson \textit{et al.} \cite{larsson2016learning} utilized hyper-column from a VGG-Net \cite{simonyan2014very} pre-trained on ImageNet for semantic feature extraction. However, it requires high computational footprints which makes the inference slower during test time. Iizuka \textit{et al.} \cite{iizuka2016let} on the other hand jointly trained a classification sub-network and auto-encoder stream. It not only obtains semantic features but also establishes a reasonable scene context for colorization. Based on a VGG-Net backbone, Zhang \textit{et al.} \cite{zhang2016colorful} introduced cross-channel encoding and class rebalancing techniques to generate unimodal distribution of color embeddings.

\begin{figure*}[t]
\centering
\includegraphics[width=\linewidth]{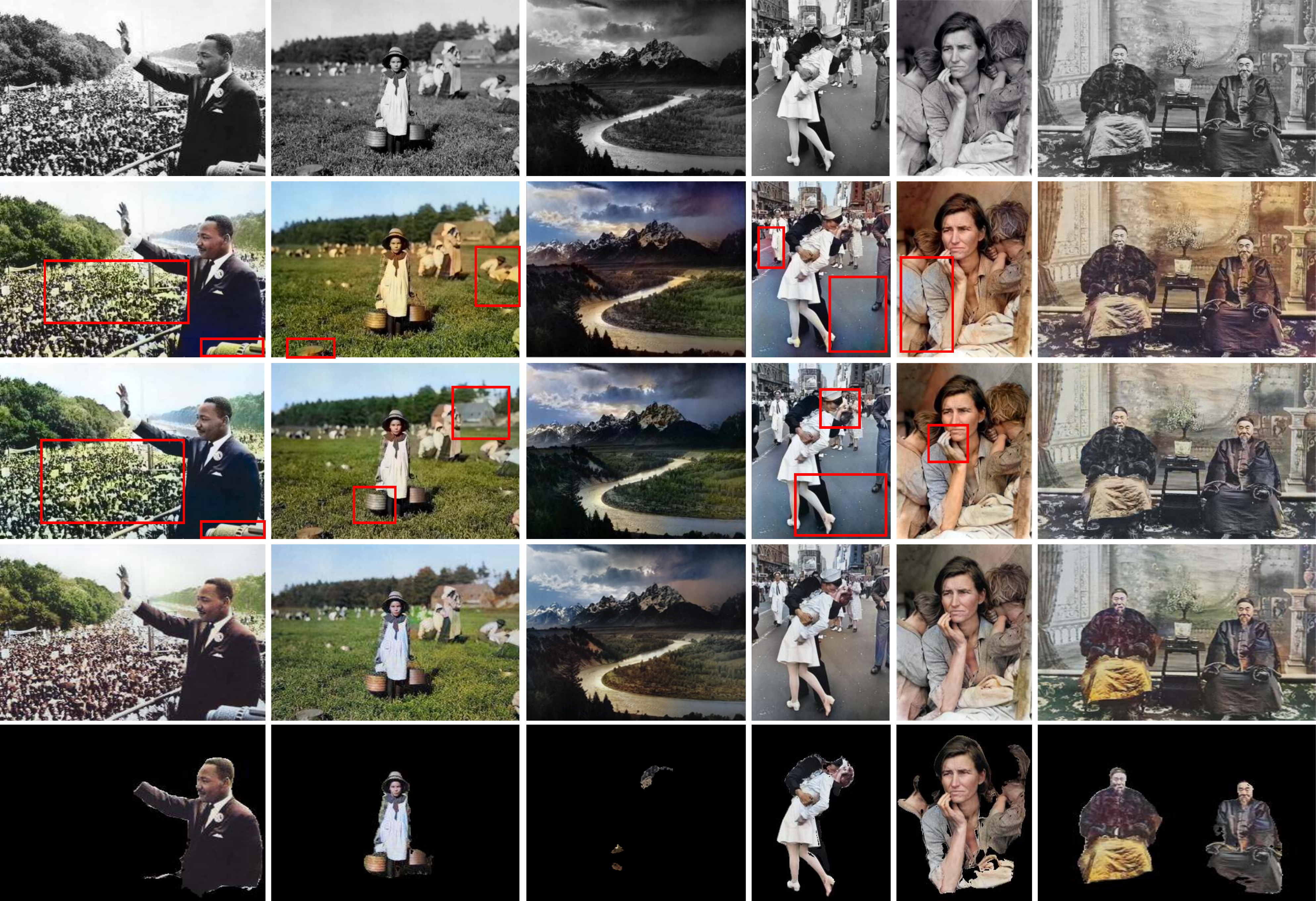}
\caption{Illustration of coloirzation results by \cite{zhang2016colorful}, \cite{vitoria2020chromagan} and proposed method on old black and white photographs. The rows from top to bottom represent grayscale input, colorization results of\cite{zhang2016colorful}, \cite{vitoria2020chromagan} and proposed method and saliency map generated by proposed method, respectively. The red rectangles highlight specific regions suffer from color bleeding or semantic confusion. Our model learns the different colorization representations in multiple scenes: speech, countryside, landscapes, city street, and human portraits. Photos were taken from the US National Archives (public domain). Please visit https://github.com/zhaoyuzhi/Semantic-Colorization-GAN (supplementary material) to see more colorization results.}
\label{demo}
\end{figure*}

The automatic image colorization systems achieve better results. However, the problems of color bleeding and unreasonable assignment of colors still exist. Figure \ref{demo} shows some common examples of the failure cases of \cite{zhang2016colorful} and \cite{vitoria2020chromagan} on some legacy photos. For instance, there is color bleeding in the first column by methods \cite{zhang2016colorful, vitoria2020chromagan} since color of trees spreads to crowd. Also, the roads and human faces are colorized in blue (in column 4 and 5, respectively) by methods \cite{zhang2016colorful, vitoria2020chromagan}. It leads to semantic confusion effect in output images. To address the problems, some regularization terms such as image gradients \cite{deshpande2017learning} and segmentations \cite{zhao2018pixel} have been added to the optimization process. However, these constrains are not useful for some situations. Since image gradients cannot represent semantics, it is hard for the colorization system to judge the colors for objects with similar boundaries, e.g. trees and crowd. In addition, only a few datasets include segmentation labels with limited categories.

Considering these limitations, we propose to use saliency map to improve the image colorization quality for following three aspects. Firstly, it identifies perceptually significant regions in the image. The colorization system can then be guided to focus more on the key objects while less influenced by the backgrounds. The key objects are richer in color while the backgrounds often contain green and blue colors, e.g. trees and sky. Moreover, it reduces the bias of the system to the colors that make up the majority of images. Secondly, it assists the network to localize objects at pixel level. It represents semantically salient areas with relatively clear boundaries. Thus, it is beneficial for colorization network to alleviate color bleeding artifact. Finally, since saliency map is adaptive to different objects in an image, it is convenient to be applied to multiple datasets in colorization area.

Specifically, we perform colorization and predict saliency map simultaneously by utilizing a Saliency Map-guided Colorization with Generative Adversarial Network (SCGAN) architecture. The proposed SCGAN has the following advantages. Firstly, it adopts dual encoders, one of which is a well-trained VGG-Net \cite{simonyan2014very} for extracting semantic information. Since semantic information is implied in this VGG-based encoder, the proposed system can distinguish plausible colors for objects with similar edges. Secondly, the decoder of proposed system has two branches for producing colorization and saliency map, respectively. To augment visual salient area, we compute multiplication of the two outputs to obtain a weighted image representing salient areas, as shown in last row of Figure \ref{demo}. Then, we leverage an attention loss to emphasize the salient areas at training. Finally, it includes two discriminators for entire image and weighted image, respectively. The adversarial training strategy \cite{goodfellow2014generative} enhances sharpness and color vividness of images. Moreover, the saliency map branch better assists the mainstream to generate plausible colorization.

In addition, conventional fully automatic colorization approaches often require large training datasets such as ImageNet \cite{russakovsky2015imagenet} and Places \cite{zhou2017places}. The proposed SCGAN can be trained on a relatively small dataset (e.g. subset of ImageNet, 0.13M images). It utilizes a saliency map-based guidance method to produce visually plausible colorization in the salient regions in the image. We notice that acquisition of large dataset in some low-level vision applications is much harder compared to natural image colorization. For example, the multispectral image colorization \cite{nyberg2018unpaired, berg2018generating, Hwang2015Multispectral} requires complex imaging system and precise alignment technique. The proposed saliency map-based guidance method is beneficial to such applications.

Compared with the existing methods, the main contributions of this paper are as follows:

1) We employ the saliency map as an additional proxy task in the proposed SCGAN that improves the performance;

2) We propose a saliency map-based guidance method that helps our system effectively predict a fine colorization with a relatively small training dataset;

3) We firstly use an effective evaluation criterion CCI (Color Colorfulness Index) to evaluate colorization quality and show its high correlation with human observers;

4) We apply SCGAN architecture with saliency map-based guidance method to multispectral image colorization and obtain state-of-the-art results.

\section{Related Work}

\textbf{Scribble-based Colorization.} Scribble-based colorization method is the most straightforward way to achieve colorization of grayscale image, but it is extremely labour-intensive. It is based on prior color scribbles and then propagates them to the rest of the image. Levin \textit{et al.} \cite{levin2004colorization} proposed an optimization-based system and assumed that the adjacent pixels with same illuminance could have similar colors. This technique was enhanced using an additional adaptive edge detection algorithm by Huang \textit{et al.} \cite{huang2005adaptive}. Yatziv \textit{et al.} \cite{yatziv2006fast} proposed a fast colorization algorithm based on the concepts of luminance-weighted chrominance blending. To enhance long-range color propagation, Xu \textit{et al.} \cite{xu2009efficient} performed an affinity-based edit scheme and Chen \textit{et al.} \cite{chen2012manifold} utilized the locally linear embedding to model the linear combination for adjacent pixels in a feature space. However, the main weakness is that they only concentrate on one aspect of propagating local or global color hints. The results are highly related to the number and location of given color scribbles. To address the ambiguity brought by sparse scribbles. Xu \textit{et al.} \cite{xu2013sparse} proposed a novel approximation scheme requiring much less time and memory and Paul \textit{et al.} \cite{paul2016spatiotemporal} proposed a 3D steerable pyramids approach for occlusion handling. Since the aforementioned methods require accurate scribbles for colorization, Zhang \textit{et al.} introduced an additional deep prior from a CNN to ensure plausible colorization when no given scribbles. Those methods are still easy to overfit to scribbles. Moreover, scribbles with similar pixel locations often lead to color bleeding in colorized images.

\textbf{Example-based Colorization.} In contrast, the example-based colorization approaches exploit color information from a reference image to guide colorization. It reduces the difficulty of choosing many color scribbles. They mainly match spatial features between reference image and input grayscale image by statistical analysis \cite{reinhard2001color, welsh2002transferring}. This idea was enhanced by characterizing the image patches using GMM \cite{tai2005local}, discriminating different regions by segmentation maps \cite{ironi2005colorization}, predicting probability for each pixel by global optimization \cite{charpiat2008automatic}, and modelling color selection by energy-minimization method \cite{bugeau2013variational}. Moreover, superpixels \cite{gupta2012image, li2017example, fang2019superpixel} were utilized to model the correspondences between grayscale input and reference. To alleviate effort of selecting proper reference images, Chia \textit{et al.} \cite{chia2011semantic} developed an image retrieval method to download appropriate reference images from Internet. However, those methods are highly based on references which are remarkably close to grayscale input.  The colors of output images often appear unnatural when given images not similar to input. In order to generalize to more reference images, He \textit{et al.} \cite{he2018deep} and Zhang \textit{et al.} \cite{zhang2019deep} applied deep image analogy technique and neural network to match the semantics of the target image and reference accurately. In addition, researchers used more types of references as guidance for colorization such as words \cite{bahng2018coloring, kim2019tag2pix} and complete sentence \cite{zou2019language}. However, the combination of examples and input grayscale image is difficult in terms of transferring examples to useful color information.

\textbf{Fully Automatic Colorization.} Recently, fully automatic colorization methods have outperformed traditional methods due to their robustness and generalization. They are based on CNN to learn mapping from grayscale to color embeddings as a self-supervised task chiefly. Cheng \textit{et al.} \cite{cheng2015deep} first adopted a deep neural network to colorize the images based on the extracted features from different patches. However, their training dataset is too small and network structure is simple. Without using handcrafted features, Larsson \textit{et al.} \cite{larsson2016learning} proposed an end-to-end CNN architecture. The hyper-column of a pre-trained VGG-Net is utilized to augment original grayscale input; whereas its memory consumption is too high. Iizuka \textit{et al.} \cite{iizuka2016let} developed a two-stream architecture to jointly predict the color embedding and category of the scene. The semantics from classification sub-network are merged into mainstream by a fusion layer. Zhang \textit{et al.} \cite{zhang2016colorful} adopted a VGG-styled network with added depth and dilated convolutions. They introduced cross-channel encoding and class rebalancing techniques to resolve the inherent ambiguity and multimodal nature of the colorization problems. However, those methods retain common artifacts in colorization area such as color bleeding and semantic confusion. To address these problems, Zhao \textit{et al.} \cite{zhao2018pixel, zhao2019pixelated} added segmentation information and Lei \textit{et al.} \cite{lei2019fully} proposed a bilateral loss for self-regularization.

Moreover, some generative models have been leveraged for multimodal colorization. Isola \textit{et al.} \cite{isola2017image} proposed a general image-to-image translation framework based on conditional GAN \cite{goodfellow2014generative}. The experimental results demonstrated that the vividness of colorized images was enhanced due to adversarial training. Deshpande \textit{et al.} \cite{deshpande2017learning} utilized a mixture density network (MDN) to map the grayscale images to GMM. There are numerous possible vectors sampled from GMM and each corresponds to a colorization type. It was enhanced by Messaoud \textit{et al.} \cite{messaoud2018structural} by introducing structural consistency. Based on capturing dependencies of neighbouring pixels to ensure color consistency, Royer \textit{et al.} \cite{royer2017probabilistic} and Sergio \textit{et al.} \cite{guadarrama2017pixcolor} developed a PixelCNN network to produce multiple plausible and vivid colorizations for a given grayscale image.

\textbf{Salient Object Detection.} The early works of salient object detection (or saliency detection) were based on hand-crafted features such as color variation \cite{cheng2014global}, boundaries \cite{yang2013saliency}, and center prior \cite{jiang2013submodular}. They preserve the edges of images well but ignore the integral structural features. To predict robust saliency maps, Li \textit{et al.} \cite{li2015visual} proposed a multiscale feature extraction for superpixel saliency detection. Liu \textit{et al.} \cite{liu2014superpixel} combined image-level and superpixel-level features into saliency detection. However, hand-crafted features are hard to generalize to different scenes. Thus, the CNN is adopted to improve generalization ability of saliency detection algorithms. Researchers developed diverse architectures such as recurrent network \cite{wang2016saliency}, encoder-decoder \cite{pan2017salgan, Zhang2018Progressive, Liu2016DHSNet, Liu2019PoolSal}, feature pyramid network \cite{wang2015deep, HouPami19Dss, Wang2019Salient, Zhao2019Pyramid, zhang2020multistage} to fuse low-level edge details and high-level semantics. Some methods \cite{Liu2016DHSNet, Zhang2018Progressive, Wang2019Salient} used attention mechanism, which further improved the accuracy due to use of dense connections for each pixel. Recently, some extensions focus on improvement of network architecture to effectively use features. For instance, Liu \textit{et al.} \cite{Liu2019PoolSal} designed a pooling-based pyramid architecture to accurately locate salient areas. Pang \textit{et al.} \cite{pang2020multi} effectively used multi-level and multi-scale information and proposed a feature aggregation module. Zhao \textit{et al.} \cite{Zhao2019Pyramid} proposed a pyramid attention network that integrates different levels of information from VGG-Net. In conclusion, edge guidance, attention mechanism and semantics greatly improve the performance of saliency detection. In this paper, we choose the approach \cite{Zhao2019Pyramid} to generate robust and accurate saliency maps.

\textbf{Generative Adversarial Network.} The GAN was proposed by Goodfellow \textit{et al.} \cite{goodfellow2014generative} to generate data in an unsupervised manner. It contains a generator that learns to produce realistic data and a discriminator that judges whether the input is generated by generator or sampled from ground truth. The system is trained to minimize the JS-divergence between generated samples and target dataset. To stabilize its convergence, some advanced divergences for estimating feature disparity were proposed, such as f-divergence \cite{nowozin2016f}, Pearson $\chi^2$ divergence \cite{mao2017least}, and Earth-Mover distance \cite{arjovsky2017wasserstein}, which was further improved by adding a gradient punishment \cite{gulrajani2017improved}. Compared to traditional pixel-level loss, the adversarial loss minimizes the various divergences between the generated images and the real images in the target domain, leading to a substantial boost of the results. The proposed SCGAN framework aims at producing perceptually high-quality colorizations.

\textbf{Comparative Analysis of Colorization Methods.} Early colorization methods often require human hints such as scribbles and reference images as guidance. They \cite{huang2005adaptive, ironi2005colorization, luan2007natural, liu2008intrinsic, gupta2012image} mainly utilized hand-crafted features including low-level SIFT or edges and high-level scene or location categories. The limitation of these works is not general to images in different scenes. Recently, deep neural networks have been utilized to address this problem.  They mainly adopted pre-trained networks to enhance colorization quality but individual optimization skills. Thus, their colorization effects are different, e.g. classifying color for pixels \cite{zhang2016colorful} promotes very colorful results; training with scene classification \cite{iizuka2016let, vitoria2020chromagan} ensures overall color correctness; contextual loss \cite{mechrez2018contextual, zhang2019deep} facilitates color similarity with ground truth. Moreover, to alleviate color bleeding and semantic confusion, additional constrains such as gradient loss \cite{deshpande2017learning}, segmentation loss \cite{zhao2018pixel, zhao2019pixelated}, and bilateral loss \cite{lei2019fully} were proposed. They worked well in some circumstances; whereas saliency map is more general to all images compared with them. In this paper, we propose to extract semantics and use a novel joint training with saliency detection.

\begin{figure*}[t]
\centering
\includegraphics[width=\linewidth]{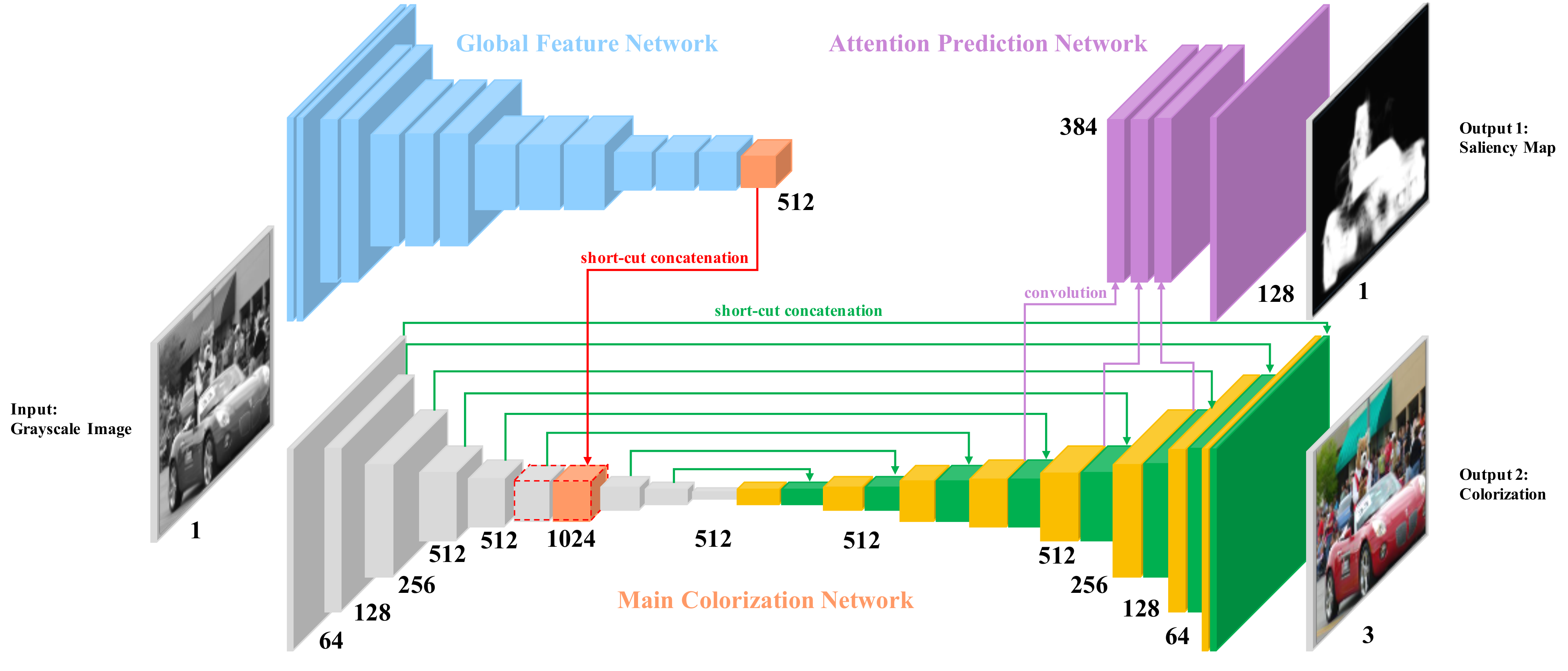}
\caption{Architecture of the proposed SCGAN generator. It receives a grayscale image as input and predicts a colorized image with a corresponding saliency map. The scalar denotes number of channels for relevant block. Different colors represent the distinct parts of generator architecture.}
\label{generator}
\end{figure*}

\begin{figure}[t]
\centering
\includegraphics[width=\linewidth]{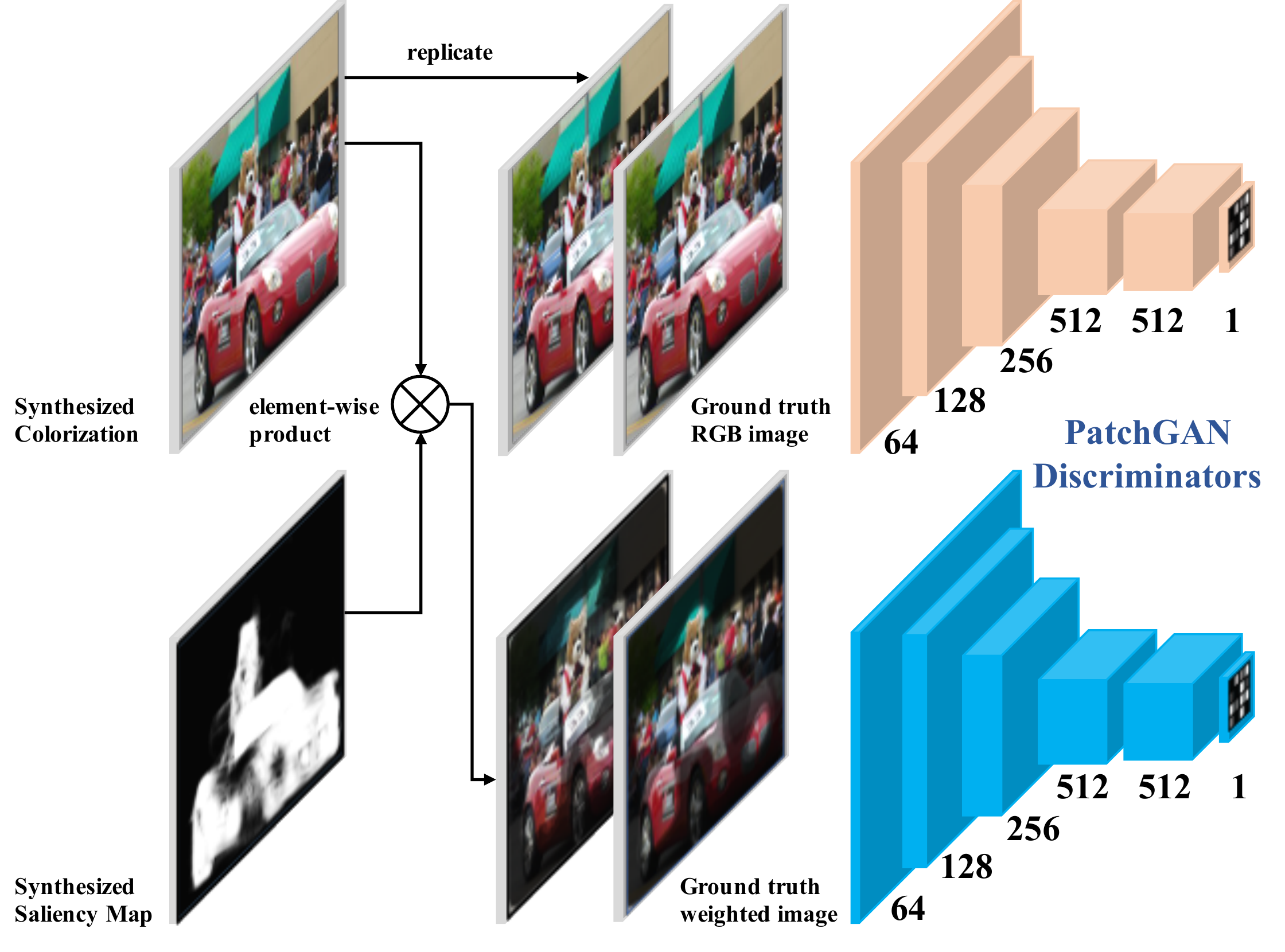}
\caption{Architecture of the proposed SCGAN discriminators. The inputs of the two discriminators (color image discriminator and attention-based discriminator) are pairs of colorized images and the images with attention region, respectively.}
\label{discriminator}
\end{figure}

\section{Methodology}

\subsection{SCGAN Architecture}

An overview of the SCGAN framework is shown in Figure \ref{generator} and Figure \ref{discriminator}. Our method is based on a hierarchical GAN architecture that produces colorizations and saliency maps from grayscale images jointly. It consists of four parts: global feature encoder, main colorization network, saliency prediction network, and patch-based discriminators. The first three components constitute generator. The global feature encoder adopts VGG-16-Gray \cite{simonyan2014very} architecture, all max-pooling layers of which are replaced by convolutional layers with stride of 2. While the main colorization network is based on U-Net structure \cite{ronneberger2015u} with skip connection between each encoder layer $i$ and decoder layer $n-i$ with same resolution, where $n$ is the total number of layers, as green lines shown in Figure \ref{generator}. It effectively prevents gradient vanishment and accelerates convergence. In order to fuse global information and local low-level information, the resultant layer of global feature encoder is concatenated with the middle layer of main colorization network. Moreover, three layers of the decoder are used to predict the saliency map with same spatial resolution as colorization.

Two discriminators share the same PatchGAN \cite{isola2017image} architecture, as shown in Figure \ref{discriminator}, which effectively models the image as Markov random field and strengthens high-frequency correctness in local image patches. The first discriminator judges the colorized image and ground truth color image. In addition, we perform element-wise product on colorized image with generated saliency map to obtain a weighted image. Similarly, we can obtain a ground truth weighted image by same computation scheme. Then, we feed the pair to the second attention-based discriminator, which judges whether the input is real weighted image or not. Based on the work in \cite{isola2017image}, we choose 70$\times$70 PatchGAN architecture including reasonable parameters for better visual quality.

\subsection{Attention Mechanism and Training Schedule}

Saliency maps are commonly used to explicitly represent the visual attention areas. Based on this observation, we assume these salient areas have more colorful patterns or higher variance, which are essential for enhancing rare colors when developing a colorization algorithm. To emphasize the areas, we propose to perform element-wise product between the colorful image and its saliency map. The output weighted image includes regions rich in color while filtering out flatten regions, as shown in Figure \ref{discriminator}. By enforcing an additional attention loss, as represented in Equation (2), on weighted image, the saliency prediction network assists the main colorization network in revising its bottom layers. Therefore, this attention mechanism serves as a kind of guidance for colorization.

Since GAN architecture is highly nonlinear, random initialization tends to converge to local minima. To facilitate and stabilize its convergence, we defined a two-phase training procedure. Firstly, SCGAN generator is only trained with L1 loss, which can remove outliers so that the generator achieves better generalization than L2 loss. Therefore, a stable adversarial training process can be maintained by striding a balance between generator and discriminators. At second stage, we construct the whole SCGAN by alternately training the generator and discriminators. Note that, the saliency map-based guidance method exists in both stages.

\subsection{Objectives}

At first stage, the L1 losses for colorized image and weighted image are jointly considered. Thus, they emphasize both pixel fidelity and perceptually significant regions of the generated images. The losses are defined as:

\begin{equation}
L_1 = \mathbb{E} [|| G_c (x) - c ||_1 ],
\label{eq_1}
\end{equation}

\begin{equation}
L_A = \mathbb{E} [|| G_c (x) \odot G_s (x) - c \odot s ||_1 ],
\label{eq_a}
\end{equation}
where $x$, $c$ and $s$ represent input grayscale image, ground truth colorful image and saliency map, respectively. The $G_c (x)$ and $G_s (x)$ are the colorized image and corresponding saliency map. The operator $\odot$ means element-wise product.

At second stage, we add two additional discriminators $D_c (*)$ and $D_A (*)$, respectively. The WGAN-GP loss \cite{arjovsky2017wasserstein} items are defined as:

\begin{equation}
L_G = - \mathbb{E} [D_c( G_c (x))] - \mathbb{E} [D_A( G_c (x)  \odot G_s (x))],
\label{eq_g}
\end{equation}

\begin{equation}
\begin{aligned}
L_D & = \mathbb{E} [D_c( G_c (x))] + \mathbb{E} [D_A( G_c (x) \odot G_s (x))] \\
    & - \mathbb{E} [D_c( c )] - \mathbb{E} [D_A( c \odot s )] \\
    & + \lambda \mathbb{E} [( || \triangledown_{\ddot c} D_c (\ddot c)||_2 - 1 )^2] \\
    & + \lambda \mathbb{E} [( || \triangledown_{\ddot s} D_s (\ddot s)||_2 - 1 )^2]
\end{aligned}
\label{eq_d}
\end{equation}
where Equation (3) and the first four terms of Equation (4) constitute original WGAN loss, the remaining of Equation (4) is a gradient penalty. According to \cite{gulrajani2017improved}, we define $\ddot c$ and $\ddot s$ sampling uniformly along straight lines between pairs of points sampled from the synthesized images $G_c (x)$, $G_s (x)$ and ground truth images $c$, $s$, respectively. The gradient penalty coefficient $\lambda$ is set to 10.

In order to improve perceptual quality, we measure the image semantic similarity in high-level feature space by perceptual loss \cite{johnson2016perceptual}. It is defined as:

\begin{equation}
L_p = \mathbb{E} [|| \phi_l ( G_c (x) ) - \phi_l ( c ) ||_1 ],
\label{eq_p}
\end{equation}
where $\phi_l (*)$ represents the features of the $l$-th layer of the pre-trained network. In our experiment, we use the ReLU \cite{nair2010rectified} activated $conv_{3\_3}$ layer of VGG-16 network pre-trained on ImageNet dataset.

The total loss function of the generator for the second stage includes Equation \ref{eq_1}, \ref{eq_a}, \ref{eq_g}, and \ref{eq_p}, which is given by:

\begin{equation}
Loss = L_1 + \lambda_G L_G + \lambda_A L_A + \lambda_p L_p.
\end{equation}

\section{Experiment}

\begin{figure*}[t]
\centering
\includegraphics[width=\linewidth]{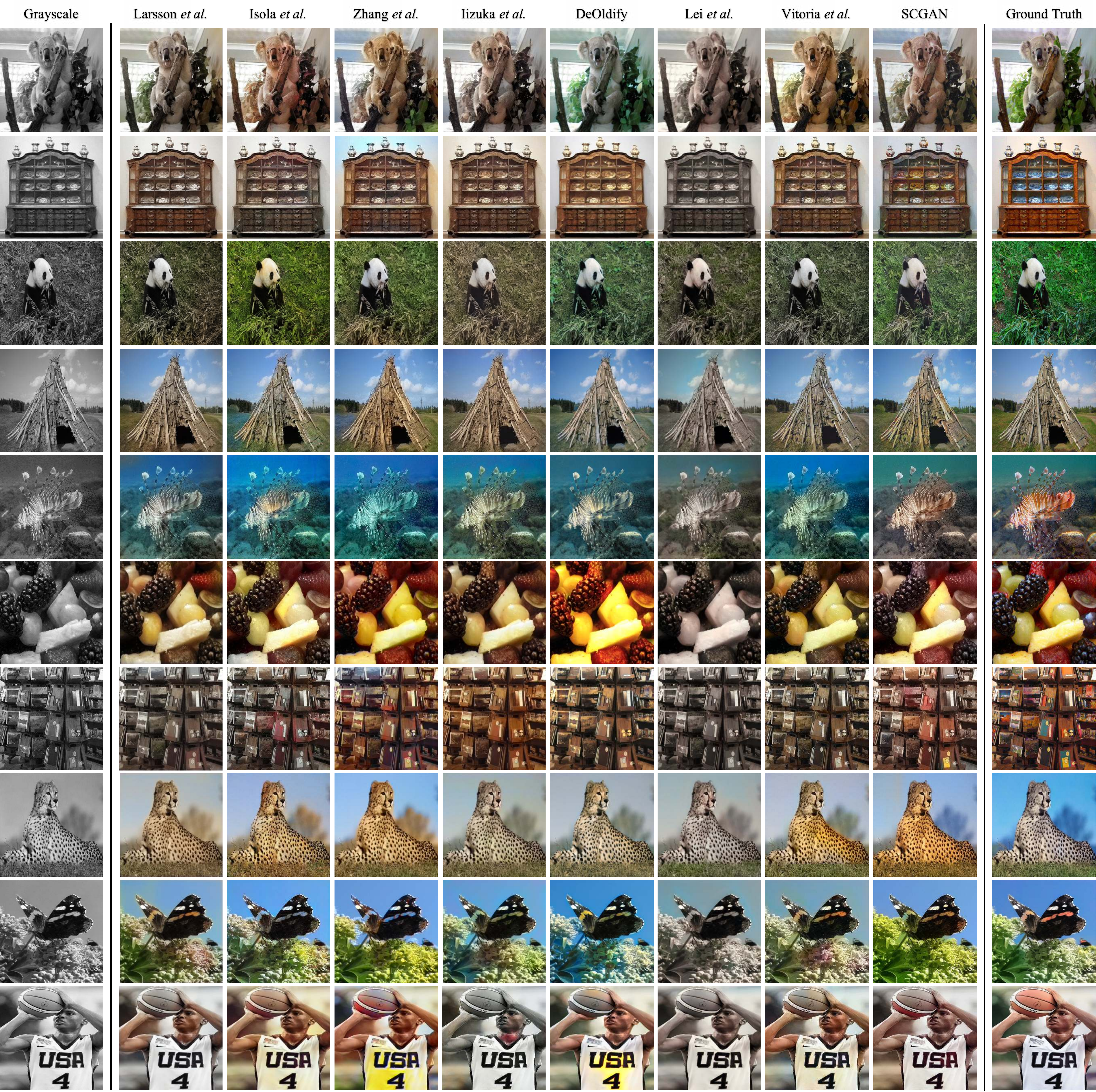}
\caption{Comparison of colorization results between the proposed SCGAN and the state-of-the-art approaches \cite{larsson2016learning, iizuka2016let, zhang2016colorful, isola2017image, DeOldify, lei2019fully, vitoria2020chromagan} by 10 samples. The first column shows the grayscale input images. Column 2-9 show automatically generated results from the state-of-the-art approaches and the proposed method. The final column shows the ground truth images.}
\label{automatic_fig}
\end{figure*}

\begin{table*}[t]
\caption{Comparison Results of SCGAN and State-of-the-art Fully Automatic Colorization Algorithms.}
\begin{center}
\begin{tabular}{|l|c|c|c|c|c|c|c|}
\hline
\textbf{Method} & \textbf{PSNR} & \textbf{SSIM} & \textbf{Top-1 Accuracy} & \textbf{CCI Ratio} & \textbf{Color Naturalness} & \textbf{Color Bleeding Removal} & \textbf{Color Colorfulness} \\
\hline
\hline
Ground Truth & / & 1 & 63.44\% & / & / & / & / \\
\hline
Grayscale & 23.23 & 0.9394 & 49.78\% & / & / & / & / \\
\hline
Larsson \textit{et al.} & \textbf{24.42} & 0.9229 & \textbf{55.16\%} & 14.93\% & 9.14 & 8.58 & 8.22 \\
\hline
Isola \textit{et al.} & 23.25 & 0.9386 & 52.29\% & 11.26\% & 8.56 & 7.93 & 8.96 \\
\hline
Zhang \textit{et al.} & 22.49 & 0.9153 & 53.97\% & 3.300\% & 9.05 & 7.10 & \textbf{9.50} \\
\hline
Iizuka \textit{et al.} & 24.32 & 0.9464 & 53.05\% & 19.60\% & 9.17 & 8.34 & 8.76 \\
\hline
DeOldify & 23.14 & 0.9194 & 53.45\% & 14.73\% & 9.20 & 8.57 & 9.01 \\
\hline
Lei \textit{et al.} & 22.96 & 0.9146 & 51.46\% & 11.40\% & 8.02 & 7.14 & 7.45 \\
\hline
Vitoria \textit{et al.} & 24.32 & 0.9273 & 53.65\% & 11.24\% & 9.03 & 8.24 & 9.17 \\
\hline
\textbf{SCGAN} & 23.80 & \textbf{0.9473} & 53.47\% & \textbf{21.41\%} & \textbf{9.32} & \textbf{8.68} & 9.04 \\
\hline
\end{tabular}
\end{center}
\label{automatic_table}
\end{table*}

\subsection{Implementation Details}

For training set, a subset of ImageNet \cite{russakovsky2015imagenet} (0.13M images) is utilized, which is only one tenth of the size of training dataset comparing with the state of the art \cite{larsson2016learning, iizuka2016let, zhang2016colorful, DeOldify, lei2019fully, vitoria2020chromagan}. We randomly sample images from 1000 categories, corresponding to the proportion of the entire dataset. It provides enough modes for SCGAN to learn the mapping robustly. The images are rescaled to 256$\times$256. They are normalized within [-1, 1] range and an additive Gaussian noise with standard deviation of 0.005 is added. In addition, the corresponding saliency maps are generated by \cite{Zhao2019Pyramid}, which are set as ground truth. They are normalized in [0, 1] range, which represent different levels of significance for salient regions.

For network architecture, the global feature network is trained from scratch until its Top-1 accuracy is verified to be sufficiently high and stable. It adopts VGG-16-Gray architecture, where each max-pooling layer is replaced with strided convolutional layer to maintain more spatial information. Batch normalization \cite{ioffe2015batch} and LeakyReLU activation function \cite{maas2013rectifier} are attached to each convolutional layer of SCGAN except the input and output layers. The reflection padding scheme is utilized to avoid border effects. Moreover, with spectral normalization \cite{miyato2018spectral} attached to each discriminator layer, the SCGAN meets 1-Lipschitz continuity.

For optimization details, the parameters of SCGAN are initialized using zero mean Gaussian distribution with standard deviation of 0.02 except global feature network. We train SCGAN generator with L1 loss and attention loss for 10 epochs at first stage and the learning rate is fixed to $2 \times 10^{-4}$. At the second stage, we train the generator and discriminators collaboratively for 30 epochs. The initial learning rate for both generator and discriminator are $1 \times 10^{-4}$ but halved every 10 epochs. We use Adam optimizer \cite{kingma2014adam} with $\beta_1$ = 0.5, $\beta_2$ = 0.999 and batch size of 8. The discriminators and generator are trained alternately until the SCGAN converges. The trade-off parameters $\lambda_G$, $\lambda_A$, and $\lambda_p$ are empirically set to 0.05, 0.5, and 5, respectively. We implement our system with PyTorch framework and train it on a NVIDIA Titan Xp GPU. It takes approximately 7 days to complete the whole training process.

\begin{figure}[t]
\centering
\includegraphics[width=\linewidth]{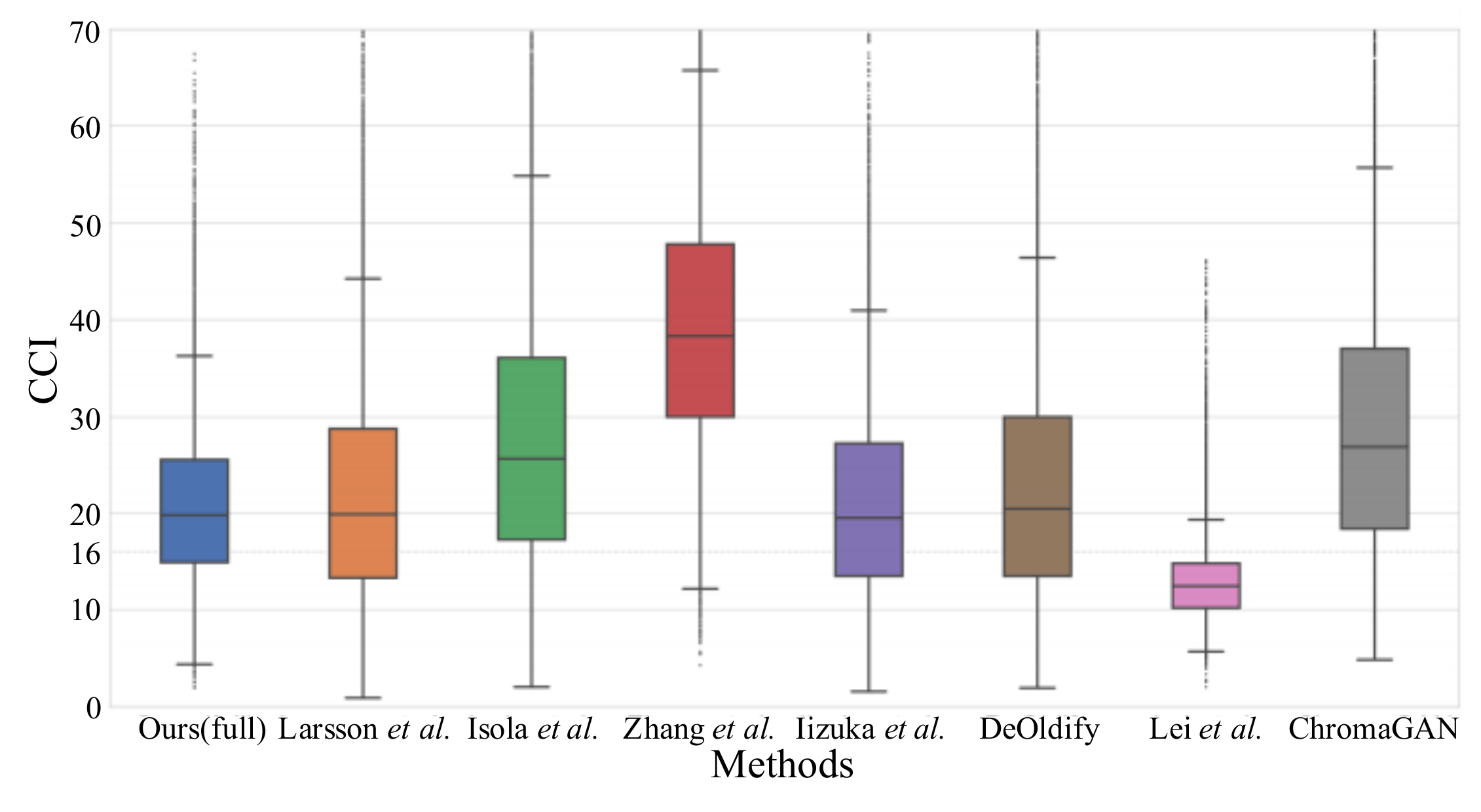}
\caption{Box plot of CCI distributions for the proposed SCGAN and state-of-the-art methods.}
\label{box}
\end{figure}

\subsection{Experimental Settings}

\textbf{Dataset.} To assess colorization quality, we set up 10000 images from ImageNet validation set \cite{russakovsky2015imagenet}, same as in \cite{larsson2016learning, zhang2016colorful} for evaluation. They are randomly selected and have a balanced representation for ImageNet categories. All the validation images encoded as grayscale are excluded and rescaled to 256$\times$256. To further demonstrate the effectiveness of several network components, we use both quantitative and qualitative analysis.

\textbf{Quantitative Metrics.} On the one hand, we apply pixel-level PSNR (peak signal to noise ratio) and SSIM (structural similarity index) \cite{wang2004image} metrics to evaluate the pixel fidelity of an image. On the other hand, since there might be many reasonable color guesses given the grayscale input, we use high-level Top-1 accuracy (computed by a well-trained VGG-16 \cite{simonyan2014very}) to measure semantic interpretability of synthesized images.

\textbf{Color Colorfulness Index.} In addition, we firstly use a new non-reference evaluator called CCI (color colorfulness index) for colorization evaluation. Basically, CCI is a professional index to measure the color vividness and naturalness \cite{hasler2003measuring, yue2018blind, huang2006natural}. Compared with traditional PSNR, CCI focuses more on color shift and saturation level. It can be viewed as a significant index for evaluating color reasonability of generated images and is defined as:

\begin{equation}
CCI_k = S_k + \sigma_k,
\end{equation}
where $S_k$ is the average saturation of image $k$, and $\sigma_k$ is the standard deviation. Note that CCI varies from 0 (achromatic image) to $\infty$ (most colorful image). However, the optimum range of CCI for a generated color image is between 16 and 20 based on large amount of experiments \cite{huang2006natural}. The correlation of optimum range of CCI with human perception equals to 95.3\%. Since the human visual system (HVS) captures color information in opponent color space \cite{yue2018blind, huang2006natural}, the RGB image is first transformed into opponent color space to compute CCI value. The transformation functions are defined as:

\begin{equation}
rg = R - G,
\end{equation}

\begin{equation}
yb = (R + G) / 2 - B.
\end{equation}

Hasler \textit{et al.} \cite{hasler2003measuring} proposed a more accurate method for computing CCI, which is used in our experiment and defined as:

\begin{equation}
CCI_k = \sigma_{rgyb} + 0.3 \mu_{rgyb},
\end{equation}

\begin{equation}
\sigma_{rgyb} = \sqrt{ \sigma_{rg}^2 + \sigma_{yb}^2 },
\end{equation}

\begin{equation}
\mu_{rgyb} = \sqrt{ \mu_{rg}^2 + \mu_{yb}^2 },
\end{equation}
where $\sigma_{*}$ and $\mu_{*}$ represent standard deviation and mean value, respectively. We calculate the ratio of the number of generated images in optimum range to the whole 10000 validation images, which represents the reasonability degree for the colorization system.

\textbf{Human Perceptual Evaluation.} Since the evaluation of color naturalness, colorfulness, and color bleeding removal are highly subjective, we perform a qualitative perceptual evaluation. The color naturalness denotes whether colorized images are similar to real-world images. It emphasizes color reasonability rather than high brightness or vividness. Conversely, color colorfulness score is high as long as generated images are very colorful, even though the color is not authentic. Moreover, color bleeding artifact exists in an image when color of one object permeates through other objects. The color bleeding removal judges the ability of colorization systems to prevent or reduce such artifact.

There are overall 20 observers participating in the test. Each observer was given 30 groups of grayscale images, ground truth colorful images, and images colorized by different algorithms. For each result, the observer was required to decide its color naturalness and severity of color bleeding by scoring 0-10. For instance, 0 represents the most achromatic or severely color bleeding images and 10 indicates the reverse. Finally, we calculate the average score across all 30 colorized images and from every observer.

\subsection{Comparison with State of the Art}

We utilize 7 state-of-the-art fully automatic algorithms \cite{larsson2016learning, iizuka2016let, zhang2016colorful, isola2017image, DeOldify, lei2019fully, vitoria2020chromagan} for comparisons. Some colorized results of proposed SCGAN and other methods are shown in Figure \ref{automatic_fig} for qualitative measurement. The results from \cite{larsson2016learning, iizuka2016let, lei2019fully} look more unsaturated than others in the second and fifth columns. In the third and fourth columns, there is semantic confusion effect. For instance, the grass of fourth row from \cite{zhang2016colorful, isola2017image} is polluted since the methods fail to classify grass and wave with similar jagged edges. Moreover, the color of sea permeates through fish, as shown in fifth row from \cite{zhang2016colorful, DeOldify, vitoria2020chromagan}. In sixth row from \cite{larsson2016learning, iizuka2016let, isola2017image, vitoria2020chromagan}, the colors of some fruits bleed into others. However, the results from proposed SCGAN are more reasonable and natural. For human perceptual evaluation, the scoring results are summarized in Table \ref{automatic_table}. The SCGAN has better performance than other methods since it produces more natural colors. The saliency map could provide attention segmentation for SCGAN, which is beneficial for removing color bleeding effect.

In addition, the results of quantitative metrics are shown in Table \ref{automatic_table}. SCGAN ranks first in the SSIM metric. It means that proposed system could accurately model the perceptual structure of reconstructed images. As PSNR is not highly related to human visual system (HVS) \cite{wang2004image}, SSIM is proposed to grasp the structural characteristics (luminance, contrast, and structure) of the image. We think SSIM is better to estimate whether the colorization is distorted or not. The proposed SCGAN with high SSIM can generate structural consistent results compared with original color images, which demonstrates the colorization is more reasonable. The SCGAN also has sound results across other quantitative evaluators.

The CCI distributions of all validation images for different algorithms are shown in Figure \ref{box}. On the one hand, methods \cite{zhang2016colorful, isola2017image} have larger CCI means and variances than others, indicating that the colorization is very saturated and color shifts very much in many generated images, respectively. Moreover, the method \cite{zhang2016colorful} has the best performance of color colorfulness but the worst color bleeding removal. It demonstrates CCI metric focuses more on color reasonability and contrast. On the other hand, methods \cite{larsson2016learning, iizuka2016let, DeOldify, vitoria2020chromagan} obtain rational CCI distribution and good PSNR and SSIM values since they generate more natural colorization. But they have less scores in perceptual evaluation and SSIM than SCGAN. Finally, the proposed SCGAN occupies the most compact distribution over CCI near the optimal range [16, 20] than other methods obviously. It demonstrates that our system produces plausible colorization and has less probability to encounter semantic confusion and color bleeding.

The human perceptual evaluation indicates that SCGAN achieves the best performance over color naturalness and color bleeding removal. Since saliency map assists the system to retain a clear separation of foreground and background, the color bleeding effect of SCGAN is less than other methods. Moreover, we use attention loss with adversarial training in SCGAN. They promote the system to strengthen colorization on key objects. Thus, SCGAN produces more natural colorizations. Zhang \textit{et al.} \cite{zhang2016colorful} obtains the highest color colorfulness due to its classification-based training scheme. From these experiment results, we notice that CCI ratio metric is highly related to color naturalness. When the average of CCI is very high, the system tends to produce colorful samples, although they may be not natural. However, it cannot represent the ability of removing color bleeding since it focuses on the global characteristic of images.

\begin{figure*}[t]
\centering
\includegraphics[width=\linewidth]{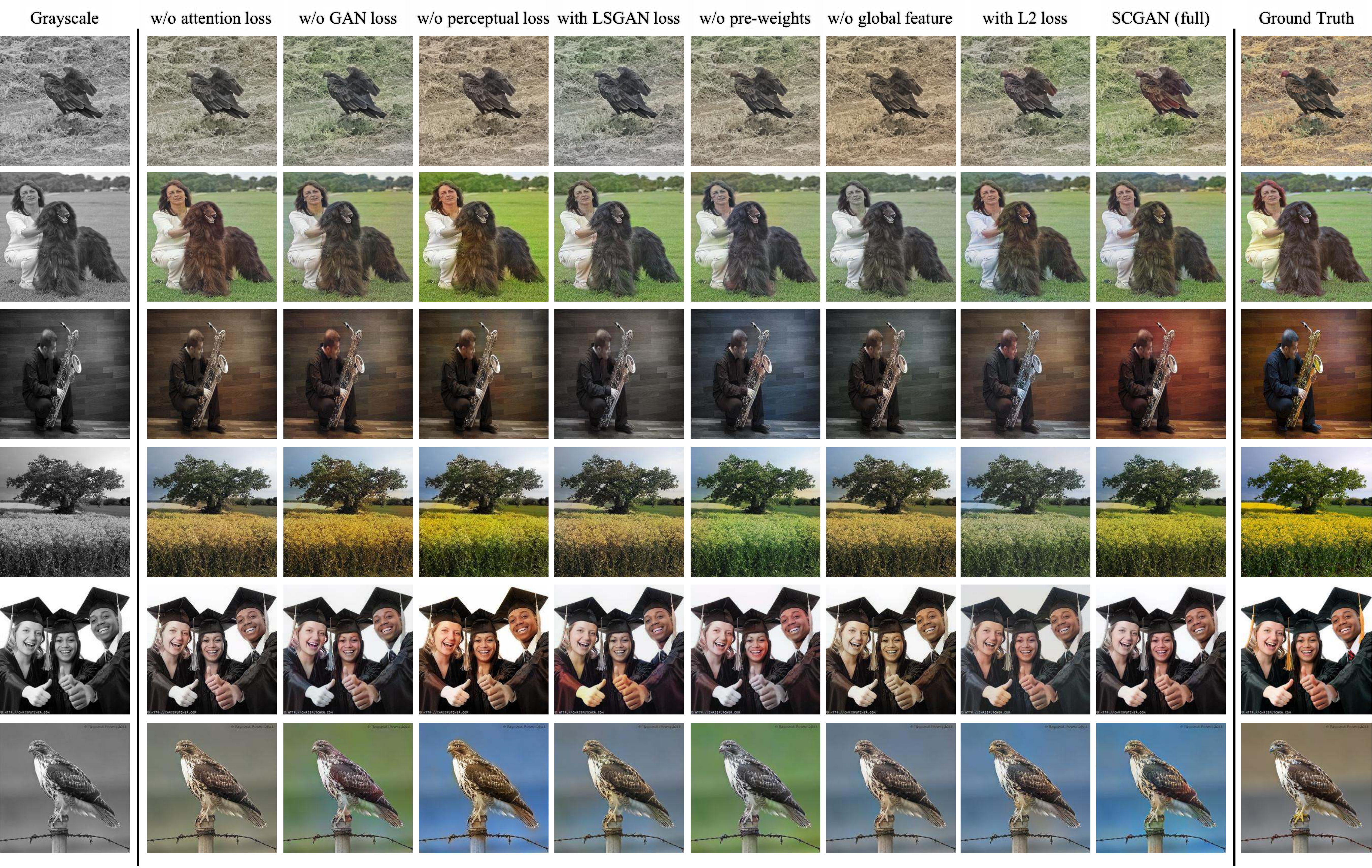}
\caption{Comparison of colorization results of different ablation study settings and full SCGAN. The first column shows the grayscale input images. Column 2-9 represent the colorization results of different settings. The final column is the ground truth colorful images.}
\label{ablation_fig}
\end{figure*}

\begin{table}[t]
\caption{Quantitative Results of Ablation Study on 10000 ImageNet Validation Set}
\begin{center}
\begin{tabular}{|l|c|c|c|c|}
\hline
\textbf{Method} & \textbf{PSNR} & \textbf{SSIM} & \textbf{Top-1 Acc} & \textbf{CCI Ratio} \\
\hline
\hline
w/o attention loss & \textbf{23.81} & 0.9368 & 52.34\% & 18.56\% \\
\hline
w/o GAN loss & 23.28 & 0.9305 & 52.89\% & 20.45\% \\
\hline
w/o perceptual loss & 23.80 & 0.9443 & 52.11\% & 21.31\% \\
\hline
with LSGAN loss & 23.46 & 0.9390 & 53.42\% & 20.86\% \\
\hline
w/o pre-weights & 23.15 & 0.9280 & 52.59\% & 18.20\% \\
\hline
w/o global feature & 23.61 & 0.9356 & 52.16\% & 17.55\% \\
\hline
with L2 loss & 23.67 & 0.9436 & 53.26\% & 19.58\% \\
\hline
\textbf{SCGAN (full)} & 23.80 & \textbf{0.9473} & \textbf{53.47\%} & \textbf{21.41\%} \\
\hline
\end{tabular}
\end{center}
\label{ablation_table}
\end{table}

\subsection{Ablation Study}

In order to further demonstrate the effectiveness of several network components, we analyze different components of our system on validation dataset quantitatively. Basically, there are 7 settings to exclude some parts from original structure:

\textbf{1) SCGAN w/o attention loss.} Drop the saliency prediction network and its corresponding discriminator in order to analyze the effect of attention mechanism of system. We utilize twice amount of data (one fifth of ImageNet training set) for training to demonstrate the effectiveness of saliency map-based guidance method;

\textbf{2) SCGAN w/o GAN loss.} Drop the two discriminators of colorized images and weighted images, with the adversarial training to analyze the adversarial loss in SCGAN. This setting will not affect the architecture of generator;

\textbf{3) SCGAN w/o perceptual loss.} Drop the perceptual loss at second stage. This setting only changes the optimization method, while the network architecture is remained;

\textbf{4) SCGAN with LSGAN loss.} Replace original WGAN-GP training strategy with LSGAN \cite{mao2017least} at the second stage. It minimizes the Pearson $\chi^2$ divergence between the generated samples and ground truth;

\textbf{5) SCGAN w/o pre-weights.} Initialize the parameters of global feature network using Gaussian distribution. It evaluates the utility of pre-trained weights for global feature network since SCGAN architecture is unchanged;

\textbf{6) SCGAN w/o global feature.} Delete the global feature network to analyze the effect of this module. Although it will reduce complexity of the system, the main idea of this setting is to evaluate the effectiveness of semantic context information;

\textbf{7) SCGAN with L2 loss.} Use L2 loss instead of L1 loss at both training stages. The optimization method remains unchanged.

As shown in Figure \ref{ablation_fig}, the full SCGAN has the best perceptual performance compared with the six ablation study settings. If global feature network or its pre-trained weights are removed, the color of generated images is unimaginative. The system without global semantics predicts wrong colorizations and causes semantic confusion. The attention loss emphasizes the significant parts, thus the main objects in colorizations are clear separated from backgrounds. For instance, the color of chicken in Figure \ref{ablation_fig} first row is obvious for full SCGAN; whereas the edges between chicken and background are blurry for system without attention loss. In addition, the perceptual loss and GAN loss enhance the sharpness of colorization. In Figure \ref{ablation_fig} column 3-5, the samples are less natural and colorful than full SCGAN.

The quantitative analysis is summarized in Table \ref{ablation_table}. Firstly, if visual saliency information and attention branch are removed (setting 1), the system tends to generate samples with shifted distribution over CCI. Although we utilize double amount of data to train the system, it lacks visual saliency information so that the optimization is altered. Moreover, we also train the system without attention loss using same data (one tenth of ImageNet training set) compared with normal process. The performance is still inferior to full losses. Secondly, GAN loss (setting 2) promotes the SCGAN to produce more colorful images. The perceptual loss (setting 3) facilitates the semantic interpretability of system and generates sharper images. L1 loss performs slightly better than L2 loss (setting 7) according to color abundance. Thirdly, the SCGAN without global feature network or its pre-trained weights (setting 5 and 6) have much worse ability to represent the semantics, leading to bad results over the metrics, especially classification accuracy. Finally, SCGAN with LSGAN loss (setting 4) produces worse results than the WGAN-GP loss. In conclusion, each component of the proposed SCGAN is indispensable.

\begin{figure}[t]
\centering
\includegraphics[width=\linewidth]{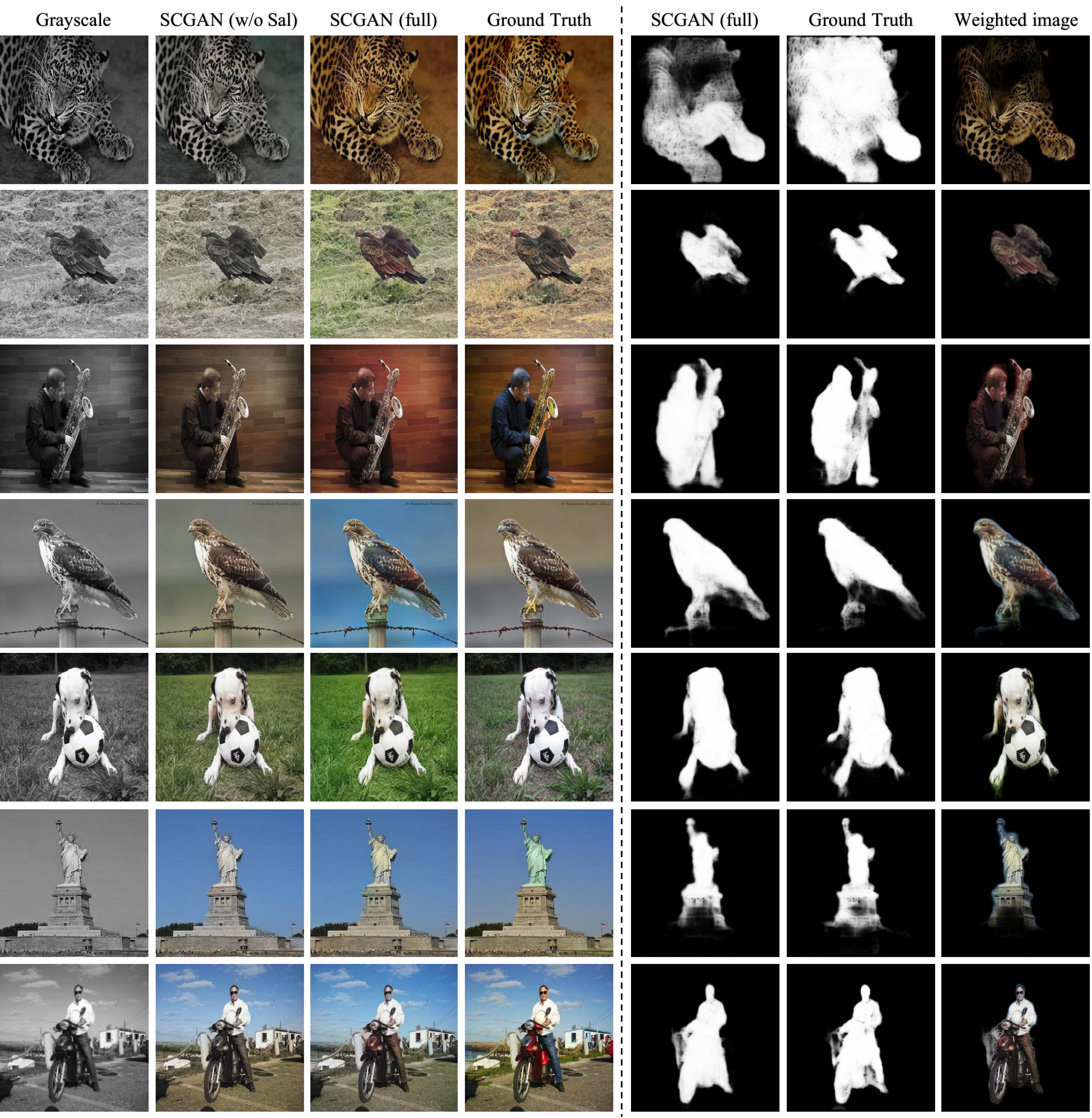}
\caption{Attention representation of the proposed SCGAN. The columns from left to right indicate that input grayscale images (1st), colorizations generated by SCGAN without saliency map (2nd), colorizations generated by full SCGAN (3rd), ground truth colorful images (4th), saliency maps generated by full SCGAN (5th), ground truth saliency maps (6th) and weighted images (7th), respectively. The weighted images are obtained by the multiplication of two outputs from SCGAN full system.}
\label{sal}
\end{figure}

\begin{figure}[t]
\centering
\includegraphics[width=\linewidth]{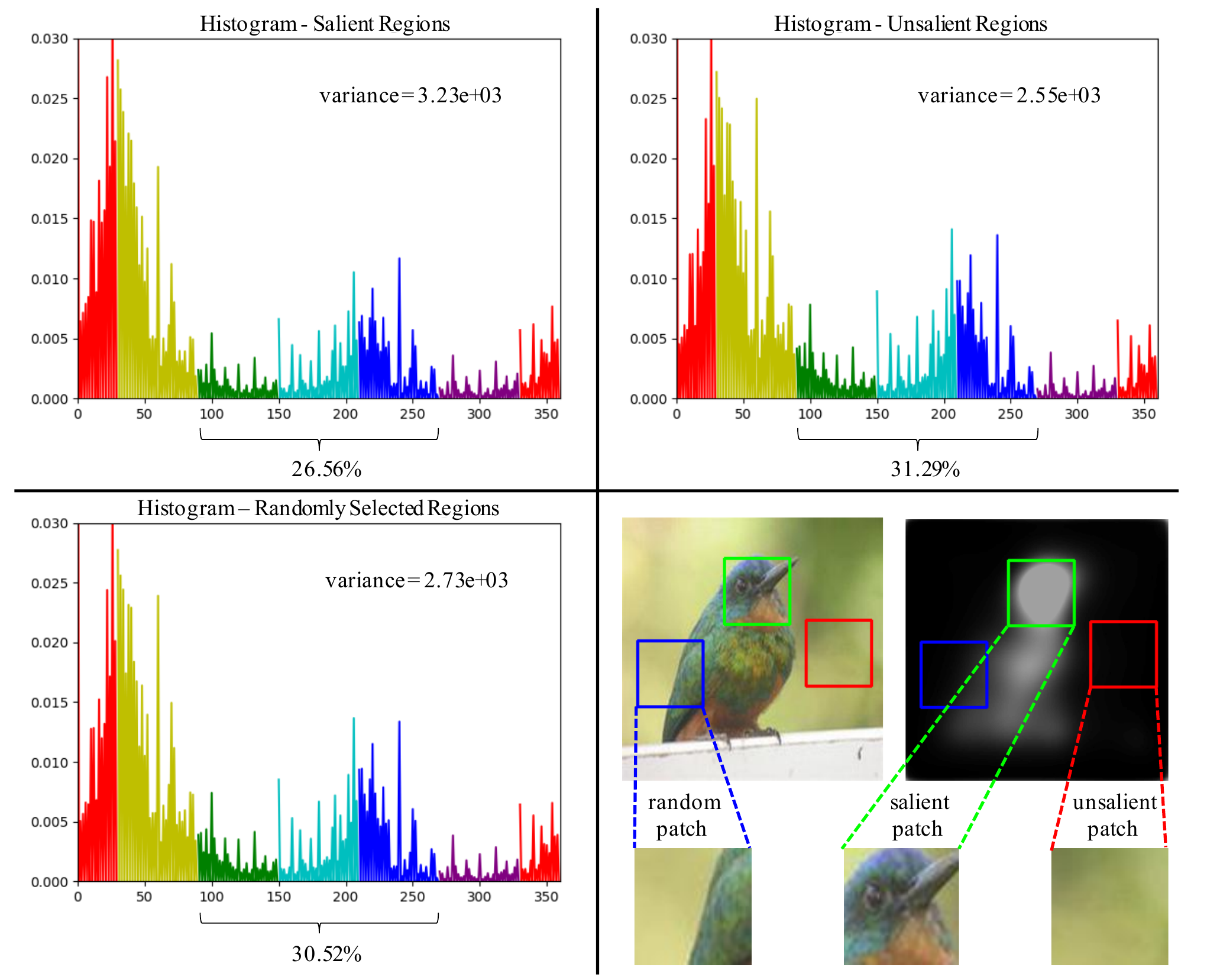}
\caption{Illustration of histogram of H channel for salient, unsalient and randomly selected regions. The percent represents the ratio of green, cyan, and blue color to all colors. The figure at lower right corner represents the patch selection scheme for the experiment. The rectangles with different colors imply 3 kinds of patches. The bird image is a training sample from ImageNet dataset.}
\label{sal2}
\end{figure}

\subsection{Saliency Map-based Guidance Method}

The SCGAN produces colorization with corresponding saliency map for grayscale image, which enhances the attention interpretation ability. As aforementioned, saliency map provides attention intensity and segmentation information \cite{wang2017saliency} in an unsupervised manner. We illustrate the attention region by the multiplication between the colorization and saliency map and comparison with SCGAN without saliency map, as shown in Figure \ref{sal}. Firstly, the foreground objects are obviously highlighted in all generated colorizations (last column of Figure \ref{sal}). The saliency maps may contribute to less color bleeding effect since the foreground and background are well separated. Secondly, the colorizations generated by full SCGAN are more colorful than it without saliency map, especially the key objects. For instance, in 1st and 4th rows, the cheetah and bird colorized by full SCGAN are more realistic. As a proxy task, the generated saliency map assists the system to pay more attention to visually important regions. This mechanism can be viewed as a guidance, making SCGAN more efficient at training stage.


In order to further demonstrate the saliency map is more efficient, we propose to compare the variance of color for salient regions and the opposite. To measure color characteristics, the H channel in HSV color space is utilized in our experiment. Since the saliency map is irregular, we alternatively choose small patches (64$\times$64 in experiment) of each training image to include high response area. The definition of salient region in RGB image is that there should be at least 80\% pixels having high value in same spatial location of corresponding saliency map. Conversely, unsalient region indicates the no response area. For fair comparison, we also count random regions as reference, as shown in Figure \ref{sal2}. The H value represents the color category, expressed by angle in HSV color space. The H variance of salient regions is much larger than unsalient regions that demonstrates they contain more colors. At training, the saliency map with attention loss stresses such regions, which contributes to the regression of diverse colors.

Moreover, the colorization system tends to learn green and blue colors first (please see supplementary for examples) since they are common in natural images, e.g. lawn and sky. The salient regions have less percent (26.56\%) of green-blue range than randomly selected regions (30.52\%) and unsalient regions (31.29\%), as shown in Figure \ref{sal2}. Thus, other colors are more probable to be utilized for SCGAN optimization. It can be regarded as a color augmentation. We believe this mechanism enhances colorization system to produce more plausible results.

\begin{figure}[t]
\centering
\includegraphics[width=\linewidth]{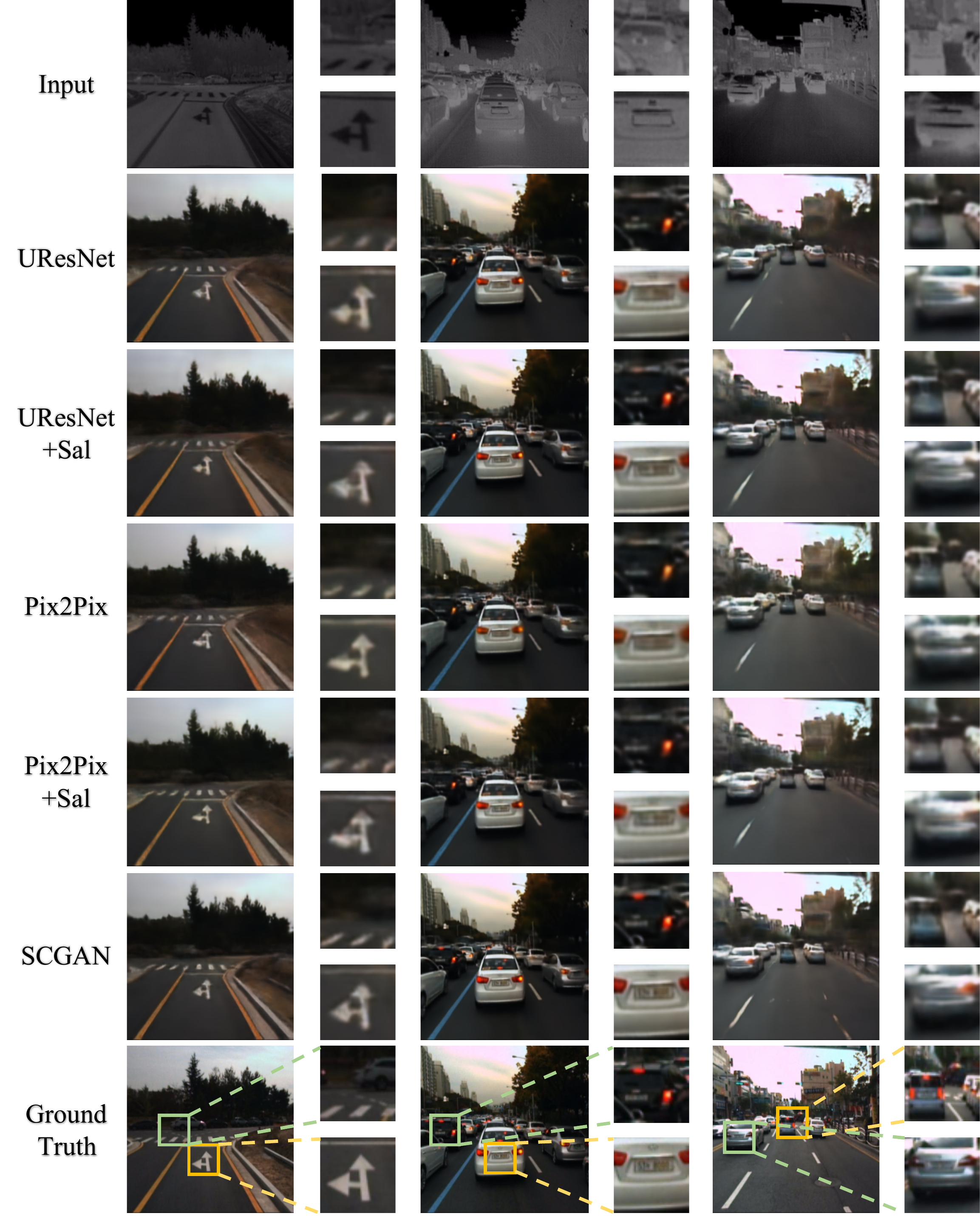}
\caption{Comparison of multispectral image colorization results between the proposed SCGAN and other approaches \cite{berg2018generating, isola2017image}. The first row is the multispectral input while last row is ground truth visible RGB images. Row 2-6 represent colorization results of proposed SCGAN and other methods. We highlight two patches in each pair and the location is indicated by green and yellow rectangles.}
\label{nir}
\end{figure}

\begin{table}[t]
\caption{Quantitative Results of multispectral Image Colorization on KAIST Validation Set}
\begin{center}
\begin{tabular}{|l|c|c|c|}
\hline
\textbf{Method} & \textbf{PSNR} & \textbf{SSIM} & \textbf{Saliency Map Guidance} \\
\hline
\hline
Pix2Pix & 23.55 & 0.8165 & - \\
\hline
Pix2Pix+Sal & 23.53 & 0.8164 & \checkmark \\
\hline
UResNet & 23.66 & 0.8219 & - \\
\hline
UResNet+Sal & 23.72 & 0.8244 & \checkmark \\
\hline
SCGAN & \textbf{24.59} & \textbf{0.8396} & \checkmark \\
\hline
\end{tabular}
\end{center}
\label{nir_table}
\end{table}

\begin{figure}[t]
\centering
\includegraphics[width=\linewidth]{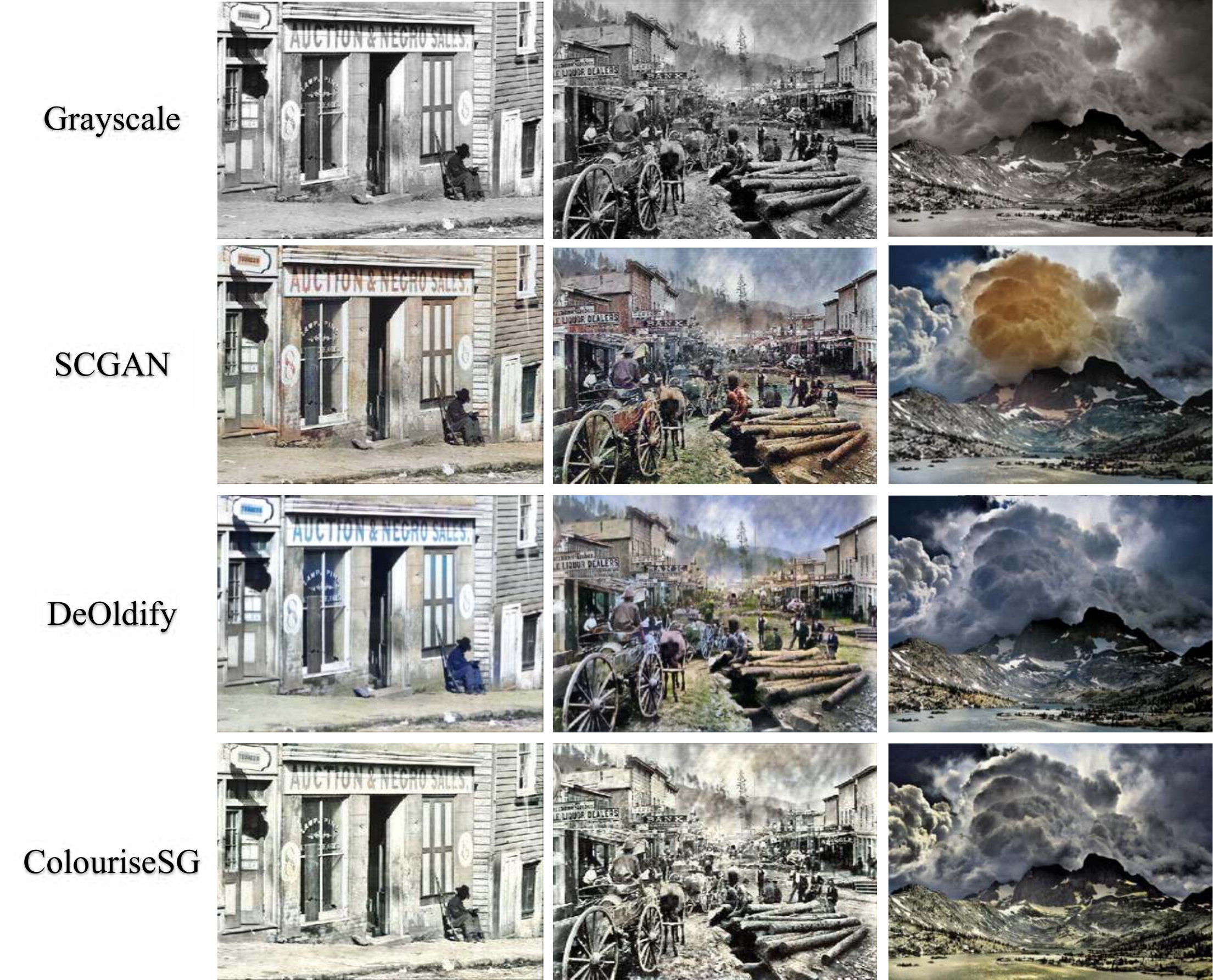}
\caption{Comparison of other legacy image-specialization colorization algorithms. The first row is the grayscale input. Row 2-4 are colorization results of the proposed SCGAN, DeOldify \cite{DeOldify}, and ColouriseSG \cite{ColouriseSG}, respectively.}
\label{legacy}
\end{figure}

\begin{figure*}[htbp]
\centering
\includegraphics[width=\linewidth]{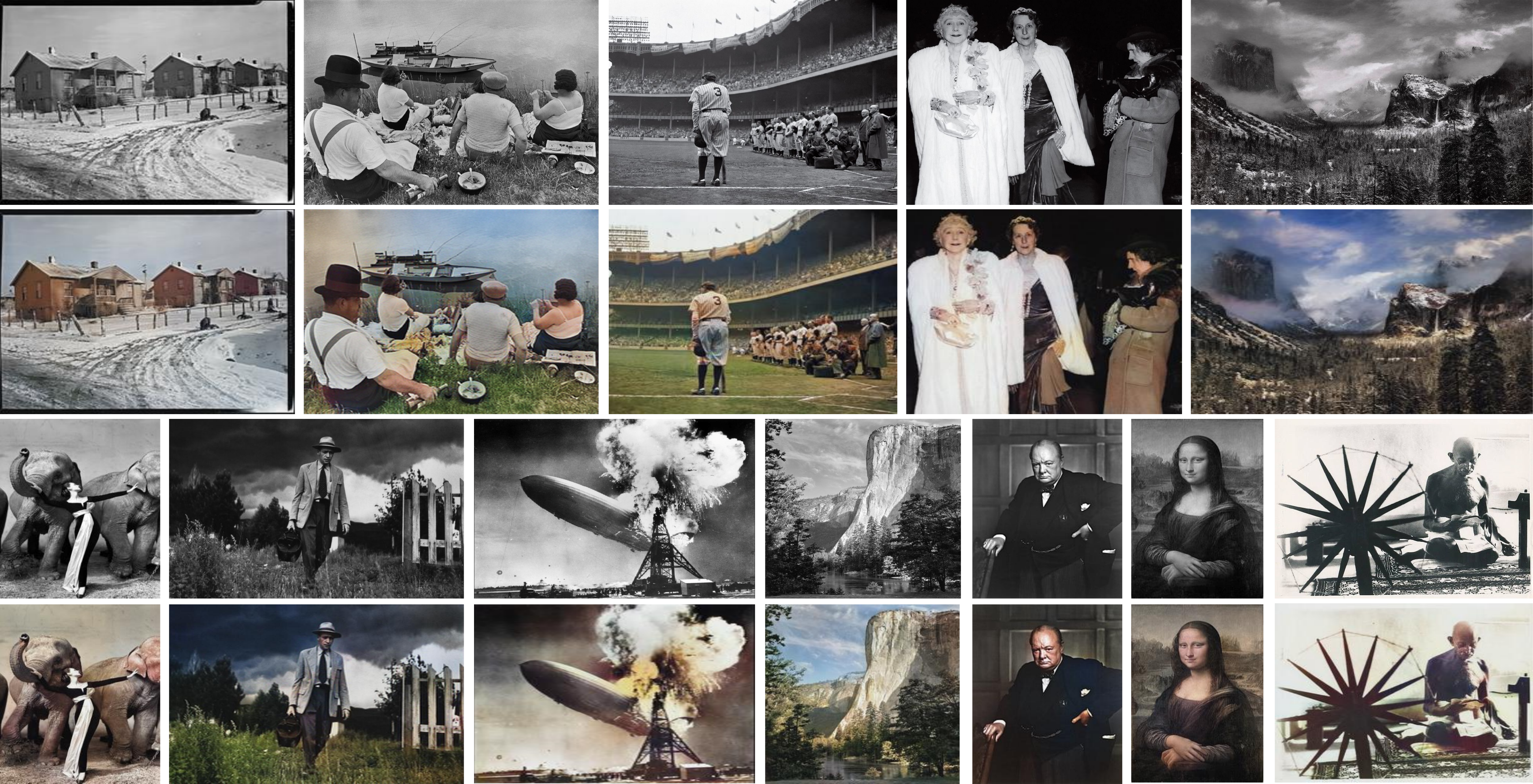}
\caption{Colorization results on historical photographs. Our colorization system still predicts visually high-quality reasonable images. The photos were taken from the US National Archives (Public Domain). For more colorized legacy photographs, please see the supplementary material.}
\label{legacy2}
\end{figure*}

\begin{figure*}[htbp]
\centering
\includegraphics[width=\linewidth]{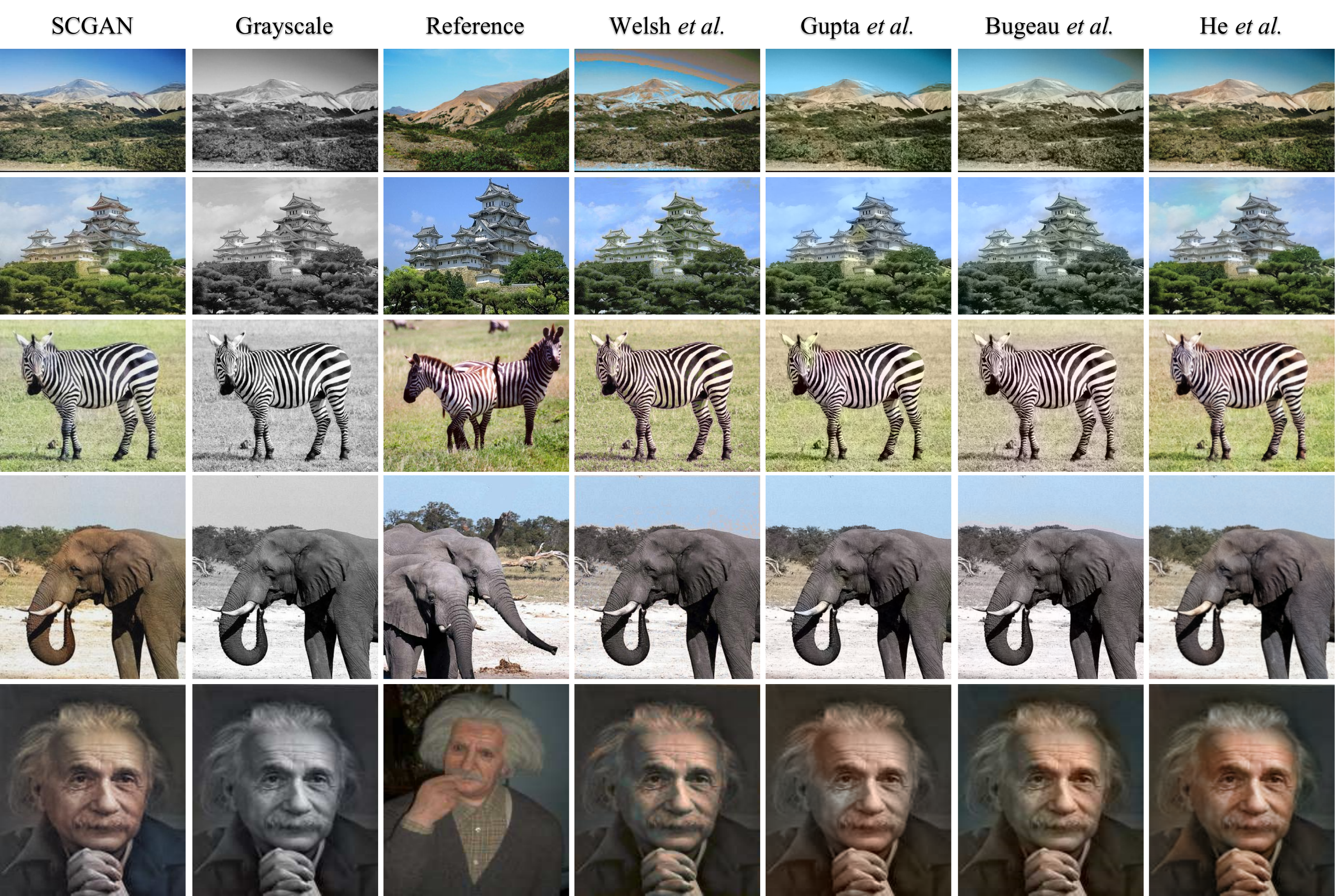}
\caption{Comparison of colorization results between the proposed SCGAN and the state-of-the-art exemplar-based algorithms \cite{bugeau2013variational, he2018deep, gupta2012image, welsh2002transferring}. The first column shows the automatic colorization results of proposed method. The second column shows the grayscale input images. The third column is the references for remaining four algorithms, which are shown in column 4-7.}
\label{example_based}
\end{figure*}

\subsection{Colorizing Multispectral Images}

In order to further verify the advance of SCGAN architecture and saliency map-based guidance method, we perform a multispectral image colorization experiment on KAIST multispectral pedestrian detection dataset \cite{Hwang2015Multispectral}. There are four network architectures used for comparison: Pix2Pix \cite{isola2017image} and it with proposed saliency map-based guidance method (Pix2Pix+Sal), UResNet \cite{berg2018generating} and it with proposed saliency map-based guidance method (UResNet+Sal), and the proposed SCGAN. The L1 loss is adopted for optimization while attention loss with same trade-off parameter $\lambda_A$ is utilized for Pix2Pix+Sal, UResNet+Sal and SCGAN. Each network is trained for 20 epochs from scratch. Some colorized results are shown in Figure \ref{nir}. If the network is trained with attention loss, the output is richer in color and clearer, e.g. the cars in Figure \ref{nir}. The results from SCGAN are sharper than other methods. Moreover, the quantitative analysis on KAIST validation set is summarized in Table \ref{nir_table}. With saliency map-based guidance method, UResNet can obtain higher PSNR and SSIM values. Since the KAIST dataset only contains approximately 90000 training pairs, which are much less than ImageNet, the function of the proposed saliency map-based guidance method is evident. The SCGAN framework achieves the best performance across all the methods, since the convolutional layers of global feature network are general to multispectral images that boost the performance. It demonstrates the SCGAN network architecture is also more advance.

\begin{figure*}[t]
\centering
\includegraphics[width=\linewidth]{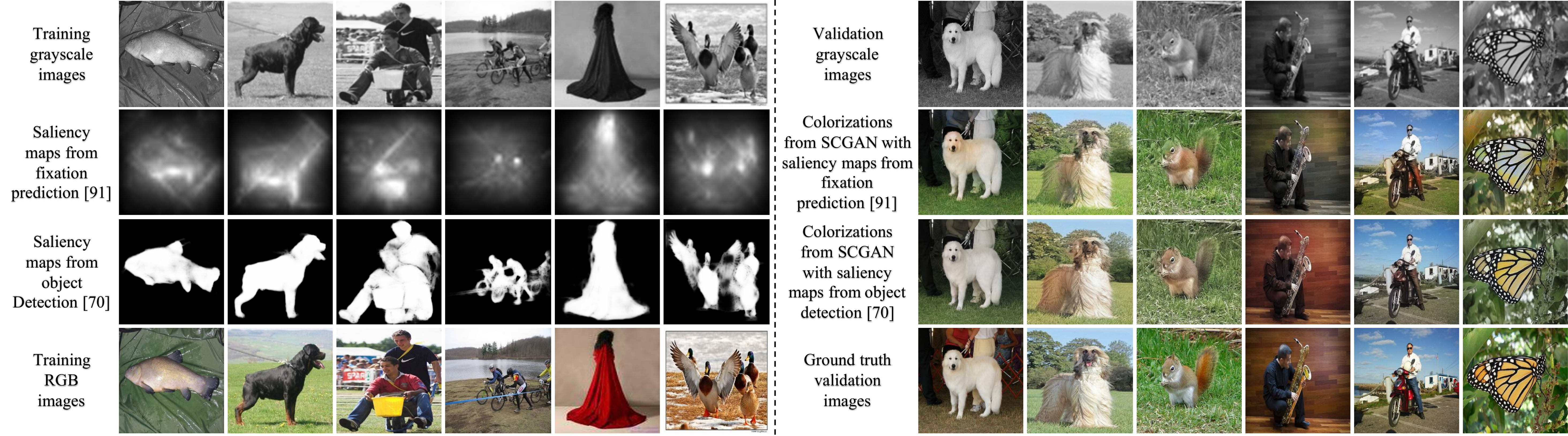}
\caption{Comparison of the two types of saliency maps. The left part of the figure includes the samples from ImageNet training set; whereas the right part represents the colorization results by SCGAN trained with different saliency maps. The first row and last row represent the grayscale input and ground truth RGB images. The saliency maps from fixation prediction and salient object detection are computed by \cite{harel2007graph} and \cite{Zhao2019Pyramid}, respectively.}
\label{twosal}
\end{figure*}

\subsection{Colorizing Legacy Photographs}

We also test SCGAN on legacy black and white photographs and illustrate the colorization results along with the results of two recent open-use automatic colorization systems \cite{DeOldify, ColouriseSG}, as shown in Figure \ref{legacy} and \ref{legacy2}. Due to the age and type of past photos and films, the statistic details are quite different from our training set and their edges are quite blur. However, SCGAN could produce plausible colorizations, which demonstrates its strong fitting ability. Since the training samples and legacy photographs are both real-world images, we further assume that SCGAN tends to learn general information primarily. During the optimization process, the system first reconstructs the profile of the objects and background. Then, it adds simple colors, like green and blue. Finally, it fixes the details and attaches special colors (please see supplementary material for illustration). It indicates that CNN-based models have strong ability to capture low-level image statistics \cite{ulyanov2018deep} while natural images have similar statistical features. It demonstrates that SCGAN has great generalization ability on legacy photographs.

\subsection{Comparison with Example-based State of the Art}

We also compare our system with state-of-the-art example-based algorithms \cite{bugeau2013variational, he2018deep, gupta2012image, welsh2002transferring}. We report the comparison results in Figure \ref{example_based}. All the test grayscale images are accompanied with corresponding references. Compared with the example-based methods, the proposed SCGAN can still generate realistic and reasonable colorizations even though there are no references. The color styles of the images generated by the proposed method are implied in the training strategy and network architecture.

\subsection{Discussion on the Usage of Saliency Maps}

Basically, there are two methods \cite{nguyen2018attentive, cong2018review} to label the ``ground truth'' saliency maps: fixation prediction \cite{itti1998model, bruce2006saliency, harel2007graph, hou2007saliency, murray2011saliency, zhang2013saliency, erdem2013visual} and salient object detection \cite{cheng2014global, yang2013saliency, jiang2013submodular, liu2014superpixel, nguyensalient, li2015visual, wang2016saliency, pan2017salgan, Zhang2018Progressive, Liu2016DHSNet, Liu2019PoolSal, wang2015deep, HouPami19Dss, Wang2019Salient, Zhao2019Pyramid, zhang2020multistage, nguyen2019semantic, pang2020multi}. The saliency maps from fixation prediction record the eye fixations of a user; whereas the saliency maps from salient object detection focus more on entire key objects. We show some saliency map samples generated by fixation prediction \cite{harel2007graph} and salient object detection \cite{Zhao2019Pyramid} in Figure \ref{twosal}, respectively. The saliency maps from salient object detection have clearer edges of objects than from fixation prediction, which are beneficial for removing color bleeding artifact. Also, the key objects have more vivid colors than other areas. To compare the effects of two types of saliency maps, we additionally train the SCGAN using the saliency maps generated from fixation prediction \cite{harel2007graph} and salient object detection \cite{Zhao2019Pyramid}, respectively. The training strategies for them are the same. Some generated saliency maps and colorization samples are shown in Figure \ref{twosal}. In first three columns of right part of Figure \ref{twosal}, there are less color bleeding artifacts for SCGAN with saliency maps from salient object detection. While in last three columns, the colorizations from row 3 are more natural than row 2. In conclusion, the SCGAN trained with saliency maps from salient object detection achieves better perceptual quality. The saliency maps in this paper denote the ones from salient object detection.

\begin{figure*}[t]
\centering
\includegraphics[width=\linewidth]{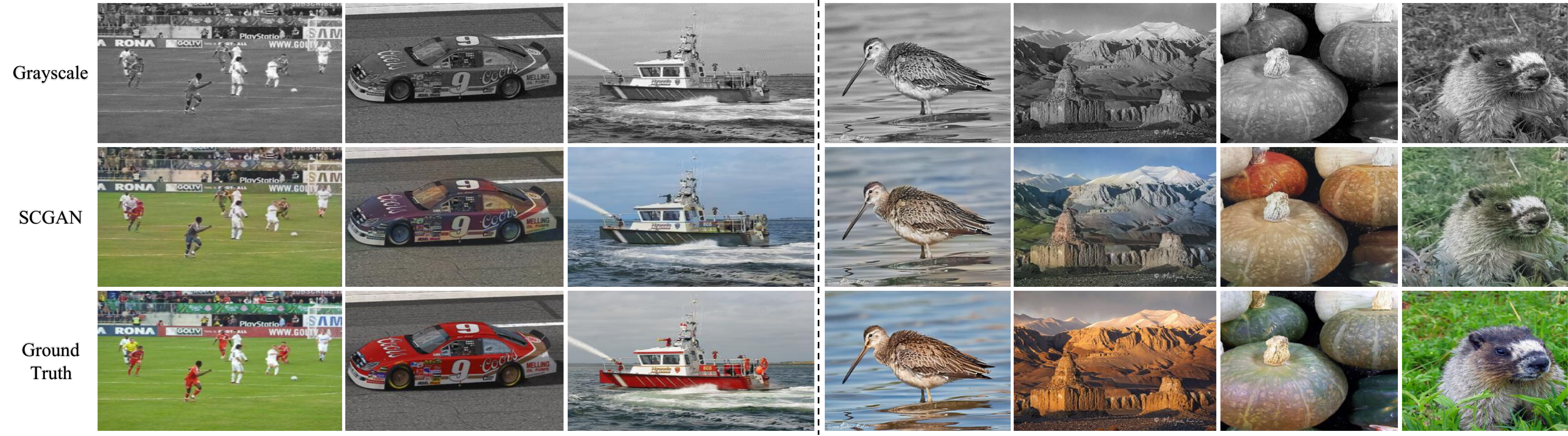}
\caption{Examples of the most common failure cases. The top, middle and bottom rows include grayscale ground truth, the generated images, and colorful ground truth, respectively. The SCGAN may be sensitive to small objects like complicated scene, special patterns, and details respectively as shown in left 3 samples. The images generated by SCGAN may be not very colorful, as illustrated in the samples.}
\label{failure}
\end{figure*}

\subsection{Failure Cases}

The proposed SCGAN can predict relatively reasonable colorizations in many samples; however, there are some common failure cases, shown in Figure \ref{failure}. It produces colorization and saliency map jointly so that core objects in images are well highlighted. There is less color bleeding effect in most of generated images. However, there is no specific loss item or network design for enhancing colors of details or small objects. Thus, SCGAN is difficult to identify plausible colors for such objects. Some failure cases are illustrated in Figure \ref{failure} first row. As we only use 0.13M training images, the system cannot include all the situations of input. Some generated images are not very colorful, as shown in Figure \ref{failure} second row. In the future, we will develop new methods generalized to small objects while generating more realistic colors.

\section{Conclusion}

In this paper, we presented a hierarchical GAN architecture called SCGAN. It generates perceptually reasonable and photorealistic colorful images and their corresponding saliency maps from grayscale input images automatically. This is achieved through a pre-trained VGG-16-Gray global feature network embedded to mainstream so that low-level and high-level semantic information are combined. In addition, we proposed a novel saliency map-based guidance method to perform the joint colorization and saliency map prediction. These designs help the system minimize semantic confusion and color bleeding in the colorized images. The proposed SCGAN framework can be trained with only one-tenth of ImageNet training data to achieve state-of-the-art colorization performance. Furthermore, we found that our system has potential to colorize multispectral images and legacy photographs with sundry scenes. Finally, we validated our system on ImageNet dataset against several state-of-the-art methods. Experiment results demonstrated that SCGAN can generate high-quality reasonable colorizations.


%

%

\section*{Acknowledgment}

We thank Mengyang Liu, Yujia Zhang, Weifeng Ou, Tiantian Zhang, Chang Zhou and Kin-Wai Lau for many helpful comments. We thank Tingyu Lin for drawing the Figure \ref{sal} and \ref{sal2}. We thank Wei Liu for the support in the revision. We also thank the anonymous reviewers and the editors for their kind suggestions.

\ifCLASSOPTIONcaptionsoff
  \newpage
\fi



%

{
\normalem
\bibliographystyle{IEEEtran}
\bibliography{bare_jrnl}
}

%
%

%

\vspace{-1cm}

\begin{IEEEbiography}[{\includegraphics[width=1in,height=1.5in,clip,keepaspectratio]{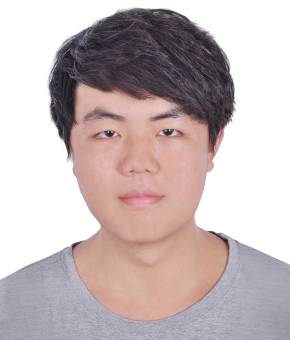}}]{Yuzhi Zhao}

(S’19) received the B.Eng. Degree in electronic information from Huazhong University of Science and Technology, Wuhan, China, in 2018. He is currently pursuing the Ph.D. degree with the Department of Electronic Engineering, City University of Hong Kong. His research interests include image processing, computational photography and deep learning.

\end{IEEEbiography}

\begin{IEEEbiography}[{\includegraphics[width=1in,height=1.5in,clip,keepaspectratio]{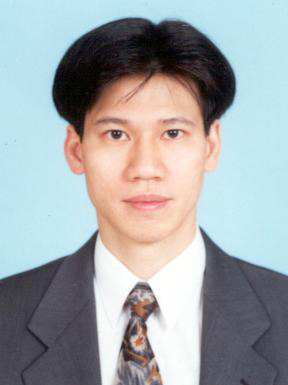}}]{Lai-Man Po}

(M’92–SM’09) received the B.S. and Ph.D. degrees in electronic engineering from the City University of Hong Kong, Hong Kong, in 1988 and 1991, respectively. He has been with the Department of Electronic Engineering, City University of Hong Kong, since 1991, where he is currently an Associate Professor of Department of Electrical Engineering. He has authored over 150 technical journal and conference papers. His research interests include image and video coding with an emphasis deep learning based computer vision algorithms.

Dr. Po is a member of the Technical Committee on Multimedia Systems and Applications and the IEEE Circuits and Systems Society. He was the Chairman of the IEEE Signal Processing Hong Kong Chapter in 2012 and 2013. He was an Associate Editor of HKIE Transactions in 2011 to 2013. He also served on the Organizing Committee, of the IEEE International Conference on Acoustics, Speech and Signal Processing in 2003, and the IEEE International Conference on Image Processing in 2010.

\end{IEEEbiography}

\vspace{-1cm}

\begin{IEEEbiography}[{\includegraphics[width=1in,height=1.5in,clip,keepaspectratio]{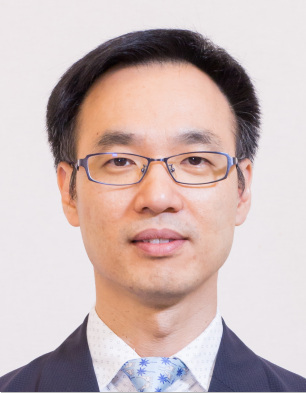}}]{Kwok-Wai Cheung}

(M’10) received the BEng, M.Sc. and Ph.D. degrees from City University of Hong Kong in 1990, 1994 and 2001, all in Electronic Engineering. He worked in Hong Kong Telecom as engineer from 1990 to 1995. He was a research assistant at the Department of Electronic Engineering, City University of Hong Kong, from 1996 to 2002. He was an Assistant Professor at Chu Hai College of Higher Education, Hong Kong from 2002 to 2016. He has been with the School of Communication, the Hang Seng University of Hong Kong as an Associate Professor since 2016. Dr. Cheung’s research interests are in the areas of image processing, machine learning and social computing.

\end{IEEEbiography}

\vspace{-1cm}

%
%

\begin{IEEEbiography}[{\includegraphics[width=1in,height=1.5in,clip,keepaspectratio]{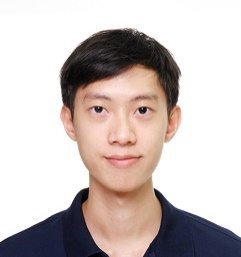}}]{Wing-Yin Yu}

received the B.Eng. degree in Information Engineering from City University of Hong Kong, in 2019. He is currently pursuing the Ph.D. degree at Department of Electrical Engineering at City University of Hong Kong. His research interests are deep learning and computer vision.

\end{IEEEbiography}

\vspace{-1cm}

\begin{IEEEbiography}[{\includegraphics[width=1in,height=2in,clip,keepaspectratio]{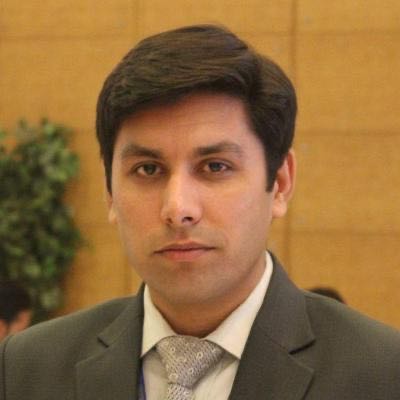}}]{Yasar Abbas Ur Rehman}

(M’19) received the B.Sc. degree in electrical engineering (telecommunication) from the City University of Science and Information Technology, Peshawar, Pakistan, in 2012, the M.Sc. degree in electrical engineering from the National University of Computer and Emerging Sciences, Pakistan, in 2015, and PhD degree in Electrical Engineering from City University of Hong Kong, Hong Kong, in 2019. He is currently working with TCL corporate research (HK) Co., Ltd as postdoctoral researcher. His research interests include the computer vision, machine learning, deep learning and its applications in facial recognition, biometric anti-spoofing, and video understanding.

\end{IEEEbiography}

%




\end{document}